\documentclass[10pt,twocolumn,letterpaper]{article}

\usepackage{cvpr}              % To produce the CAMERA-READY version
\usepackage{graphicx}
\usepackage{amsmath}
\usepackage{amssymb}
\usepackage{booktabs}
\usepackage[accsupp]{axessibility}  % Improves PDF readability for those with disabilities.
\usepackage{multirow}
\usepackage{caption}

\usepackage[pagebackref,breaklinks,colorlinks]{hyperref}

\usepackage[capitalize]{cleveref}
\crefname{section}{Sec.}{Secs.}
\Crefname{section}{Section}{Sections}
\Crefname{table}{Table}{Tables}
\crefname{table}{Tab.}{Tabs.}

\def\confName{CVPR}
\def\confYear{2024}

\makeatletter
\def\thanks#1{\protected@xdef\@thanks{\@thanks
        \protect\footnotetext{#1}}}
\makeatother

\begin{document}

%%%%%%%%% TITLE - PLEASE UPDATE
\title{AIGIQA-20K: A Large Database for AI-Generated Image Quality Assessment}

% \author{Chunyi Li$^{1}$$^{*}$, Tengchuan Kou$^{1}$$^{*}$, Yixuan Gao$^{1}$, Yuqin Cao$^{1}$, Wei Sun$^{1}$, Zicheng Zhang$^{1}$, Yingjie Zhou$^{1}$\\Zhichao Zhang$^{1}$, Weixia Zhang$^{2}$,  Haoning Wu$^{3}$, Xiaohong Liu$^{4,\dag}$, Xiongkuo Min$^{1,\dag}$, Guangtao Zhai$^{1,2,\dag}$\\
% Institute of Image Communication and Network Engineering, Shanghai Jiao Tong University$^{1}$\\
% MoE Key Lab of Artificial Intelligence, Shanghai Jiao Tong University$^{2}$\\
% S-Lab for Advanced Intelligence, Nanyang Technological University$^{3}$\\
% John Hopcroft Center, Shanghai Jiao Tong University$^{4}$\\
% {\tt\small \{lcysyzxdxc,2213889087,gaoyixuan,caoyuqin,sunguwei,zzc1998,zyj2000,liquortect\}@sjtu.edu.cn}\\
% {\tt\small haoning001@e.ntu.edu.sg, \{zwx8981,xiaohongliu,minxiongkuo,zhaiguangtao\}@sjtu.edu.cn}
% \thanks{$^{*}$ Equal contribution.}
% \thanks{$^{\dag}$ Corresponding authors.}}
\author{Chunyi Li$^{1}$$^{*}$, Tengchuan Kou$^{1}$$^{*}$, Yixuan Gao$^{1}$, Yuqin Cao$^{1}$, Wei Sun$^{1}$, \\ Zicheng Zhang$^{1}$, Yingjie Zhou$^{1}$, Zhichao Zhang$^{1}$, Weixia Zhang$^{1}$,  Haoning Wu$^{2}$, \\Xiaohong Liu$^{1,\dag}$, Xiongkuo Min$^{1,\dag}$, Guangtao Zhai$^{1,\dag}$\\
Shanghai Jiao Tong University$^{1}$\\
Nanyang Technological University$^{2}$\\
% \author{Anonymous Author \\ \\ \\
% Anonymous Address \\
%  \\
% {\tt\small \{lcysyzxdxc,2213889087,gaoyixuan,caoyuqin,sunguwei,zzc1998,zyj2000,liquortect\}@sjtu.edu.cn}\\
% {\tt\small haoning001@e.ntu.edu.sg, \{zwx8981,xiaohongliu,minxiongkuo,zhaiguangtao\}@sjtu.edu.cn}
\thanks{$^{*}$ Equal contribution.}
\thanks{$^{\dag}$ Corresponding authors.}}

\maketitle
%%%%%%%%% ABSTRACT
\begin{abstract}
With the rapid advancements in AI-Generated Content (AIGC), AI-Generated Images (AIGIs) have been widely applied in entertainment, education, and social media. However, due to the significant variance in quality among different AIGIs, there is an urgent need for models that consistently match human subjective ratings. To address this issue, we organized a challenge towards AIGC quality assessment on NTIRE 2024 that extensively considers 15 popular generative models, utilizing dynamic hyper-parameters (including classifier-free guidance, iteration epochs, and output image resolution), and gather subjective scores that consider perceptual quality and text-to-image alignment altogether comprehensively involving 21 subjects. This approach culminates in the creation of the largest fine-grained AIGI subjective quality database to date with 20,000 AIGIs and 420,000 subjective ratings, known as AIGIQA-20K. Furthermore, we conduct benchmark experiments on this database to assess the correspondence between 16 mainstream AIGI quality models and human perception. We anticipate that this large-scale quality database will inspire robust quality indicators for AIGIs and propel the evolution of AIGC for vision. The database is released on 
% \url{Anonymous link}.
\url{https://www.modelscope.cn/datasets/lcysyzxdxc/AIGCQA-30K-Image}.

\end{abstract}

\begin{figure}[tbph]
    \centering
    \includegraphics[width = \linewidth]{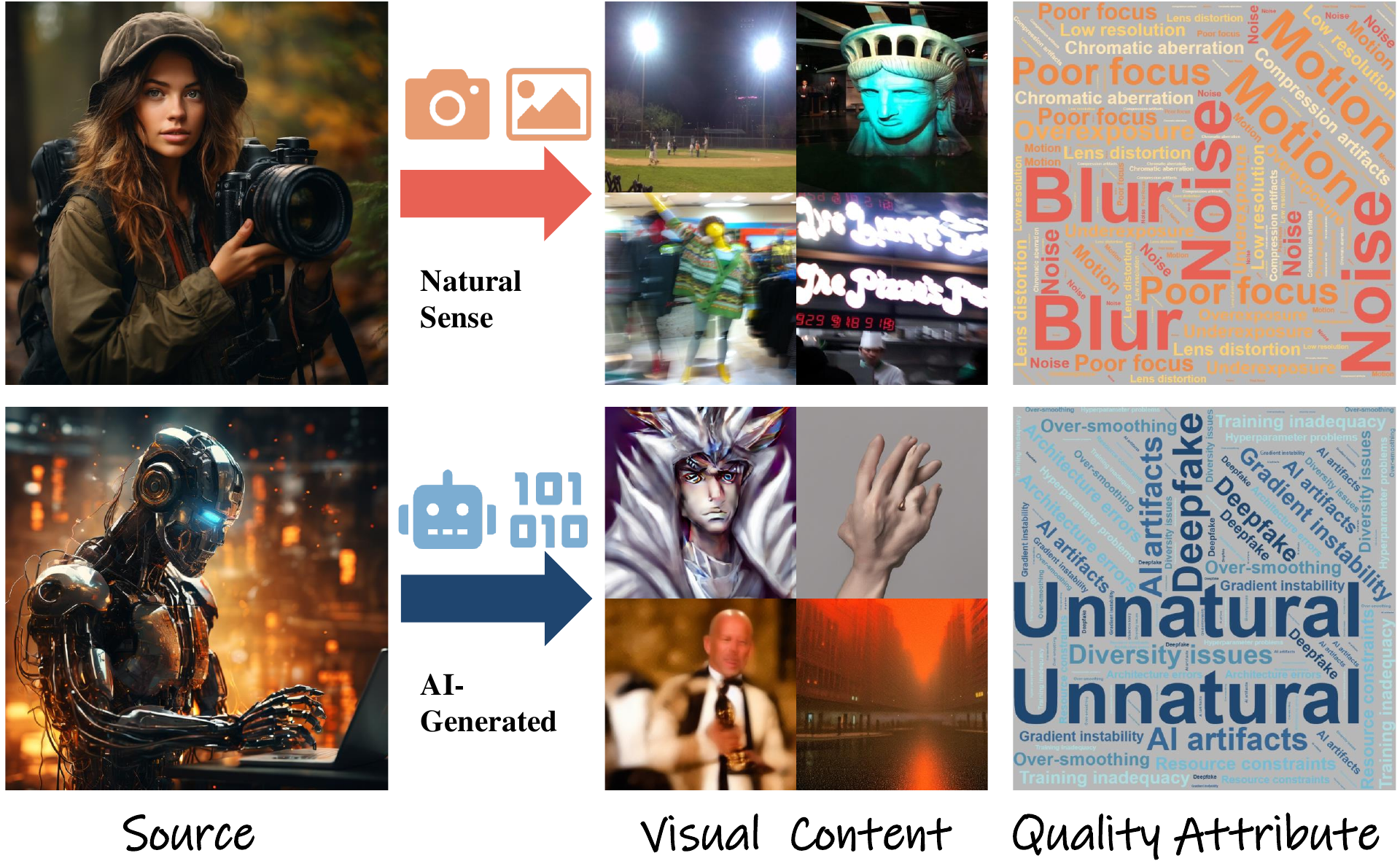}
    \caption{Illustration of the difference between Natural Sense Content and AI-Generated Content, whose perceptual quality are affected by different attributes.}
    \label{fig:spotlight}
\end{figure}
%%%%%%%%% BODY TEXT
\section{Introduction}
\label{sec:intro}
\underline{AI} \underline{G}enerated \underline{C}ontent (\textbf{AIGC}) refers to various types of content generated by artificial intelligence, such as images, videos, texts, and music. Among those modalities, AI-Generated Images (AIGIs), especially Text-to-Image (T2I) models, have already revolutionized the paradigm of entertainment, education, and social media. According to huggingface\footnote{https://huggingface.co, data collected in March 2024}, there were 10,000+ T2I models coexisting on the internet that generated results of widely varying quality.
As vision is the dominant way for humans to perceive external information, a universal quality indicator for this new visual information is a topic worth investigating in the AIGC era.
\begin{table*}[tbph]
\centering
\caption{Existing quality databases for AI-Generated Images/Videos.}
\label{tab:relate}
\begin{tabular}{cccccccc}
\toprule
\textbf{Database} & Grain          & Size  & Ratings  & Models & CFG & Iteration & Resolution      \\ \midrule
HPD\cite{database/align:HPS}               & Coarse-grained & 98,807  & 98,807  & 1      & Fixed & Fixed & Fixed        \\
ImageReward\cite{database/align:ImageReward}       & Coarse-grained & 136,892 & 136,892 & 3      & Fixed & Fixed & Dynamic         \\
Pick-A-Pic\cite{database/align:PickAPic}        & Coarse-grained & 500,000 & 500,000 & 6      & Fixed & Fixed & Dynamic        \\
AGIQA-1K\cite{database:agiqa1k}          & Fine-grained   & 1,080   & 23,760  & 2      & Fixed & Fixed & Fixed         \\
AGIQA-3K\cite{database:agiqa3k}          & Fine-grained   & 2,982   & 125,244  & 6      & Dynamic & Dynamic & Fixed         \\
AIGCIQA\cite{database:aigciqa}           & Fine-grained   & 2,400   & 48,000  & 6      & Fixed & Fixed & Fixed         \\
AGIN\cite{database:aigciqa}              & Fine-grained   & 6,049   & 181,470 & 18     & Dynamic & Fixed & Fixed         \\
% VBench\cite{database:vbench}            & Coarse-grained & 6,984   & 6,984   & 4      & Fixed & Fixed &video         \\
% FETV\cite{database:fetv}              & Fine-grained   & 2,476   & 22,284  & 4      & Dynamic & Fixed &video         \\
AIGIQA-20K        & Fine-grained   & 20,000  & 420,000 & 15     & Dynamic & Dynamic & Dynamic \\ \bottomrule
\end{tabular}
\end{table*}

However, existing Image Quality Assessment (IQA) metrics ~\cite{my:aspect-qoe,my:cartoon,my:xgc-vqa} cannot be applied in AIGIs directly. As Figure \ref{fig:spotlight} shows, the quality of Natural Sense Images/Videos (NSIs) is determined by distortion in the imaging process (e.g. blur, noise) ~\cite{add:5G,add:6G} while the quality of AIGIs is more closely related to hardware limitations and technical proficiency ~\cite{add:q-bench,add:q-instruct,add:q-boost,add:q-refine,add:openended} (e.g. unnatural, deepfake). Besides, T2I alignment is also an important factor for AIGI which is absent in traditional IQA tasks. The quality of AIGC is a mixture of perceptual and alignment quality. Therefore, towards a strong quality indicator specifically for AIGC, an AIGI quality database is highly demanded to illustrate their quality-aware attributes besides NSIs.

In the past year, the emerging demand for AIGI quality has spawned several related databases as shown in Table \ref{tab:relate} including two main categories: coarse-grained and fine-grained. 
The former usually has a larger data size, with only one user scoring the images or selecting preferences for image pairs. Thus, such scoring has strong discontinuities and bias; the latter has a smaller scale, but the quality scores are derived from the Mean Opinion Score (MOS) ~\cite{add:dh-rr,add:dh-gms,add:dh-advancing} of more than 15 users, which accurately characterizes the image quality. Meanwhile, along with the rapid development of generative models, the AIGI quality database needs to consider an increasing number of models. Besides, the quality of AIGI not only depends on the T2I model itself, where the hyper-parameters also play a decisive role. Thus, to reflect the actual distortion of AIGI, these factors also need to be dynamically adjusted.

Facing the above challenges, this paper lays the foundation of the NTIRE 2024 AIGCQA Grand Challenge to inspire effective quality metrics for AIGC, which contributes ({\romannumeral 1}) a quality database named AIGIQA-20K that extensively covers 15 T2I models. Meanwhile, it dynamically adjusts for both resolution and hyper-parameters for the first time, which comprehensively characterizes the visual distortion of AIGC. ({\romannumeral 2}) a comprehensive set of subjective quality labels. For the AIGIQA-20K, we organized 21 subjects to produce accurate MOS scores. As a fine-grained AIGI database, it has the largest size to date. ({\romannumeral 3}) an exhaustive benchmark experiment for AIGC quality assessment. The indicators cover both traditional IQA \cite{add:stable,add:light} and T2I alignment methods, which can inspire more accurate quality metrics in the future.
The rest of the paper is organized as follows. In Sec \ref{sec:construction}, details of the proposed AIGIQA-20K are provided. 
Sec \ref{sec:subjective} analyze the features of AIGIs subjective scoring. 
Sec \ref{sec:experiment} validate the several quality indicators on the AIGIs.
Finally, a conclusion is provided in Sec \ref{sec:conclusion}.

\begin{figure*}[t]
\begin{minipage}[(b) ]{0.33\linewidth}
  \centering
  \centerline{\includegraphics[width = \textwidth]{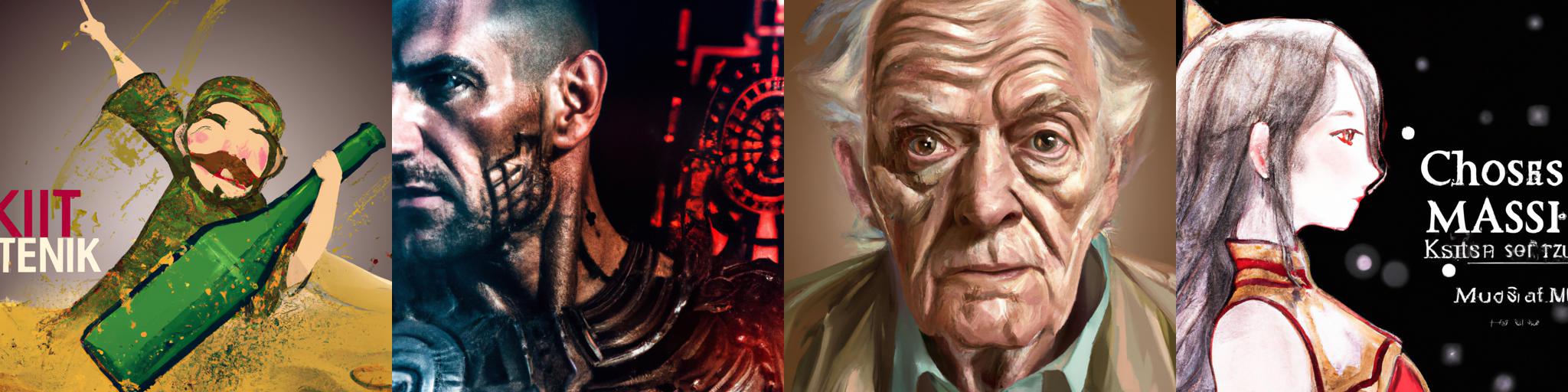}}
  \centerline{DALLE 2 \cite{gen:dalle}}\medskip
  \end{minipage}
\begin{minipage}[(b) ]{0.33\linewidth}
  \centering
  \centerline{\includegraphics[width = \textwidth]{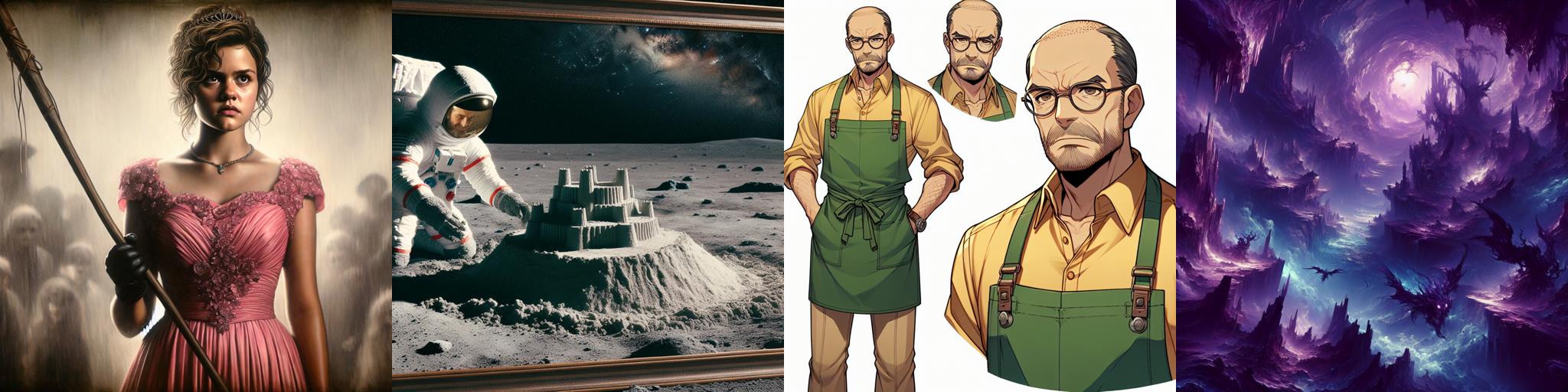}}
  \centerline{DALLE 3 \cite{gen:dalle}}\medskip
\end{minipage}
\begin{minipage}[(b) ]{0.33\linewidth}
  \centering
  \centerline{\includegraphics[width = \textwidth]{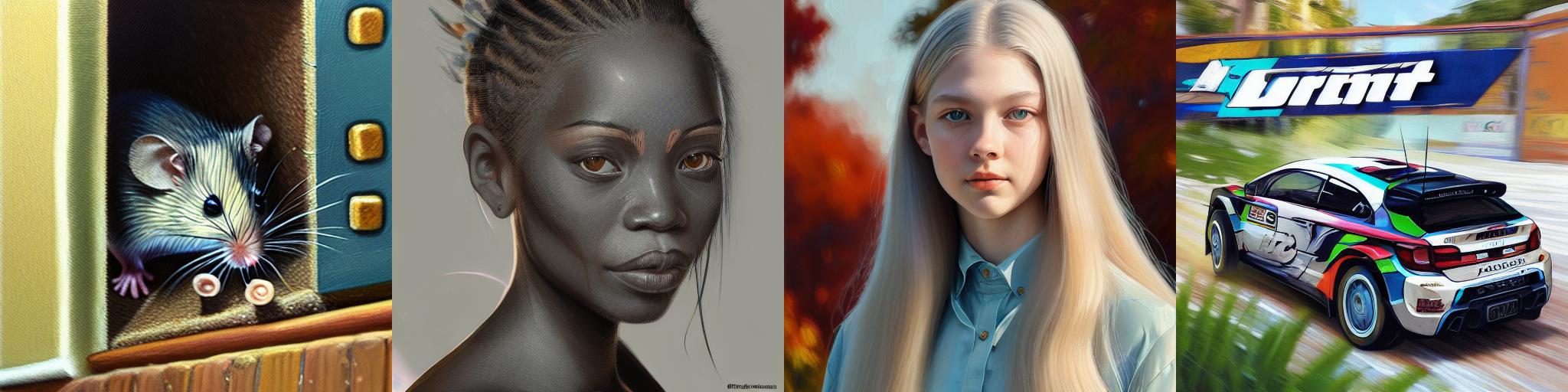}}
  \centerline{Dreamlike \cite{gen:dream}}\medskip
  \end{minipage}
\begin{minipage}[(b) ]{0.33\linewidth}
  \centering
  \centerline{\includegraphics[width = \textwidth]{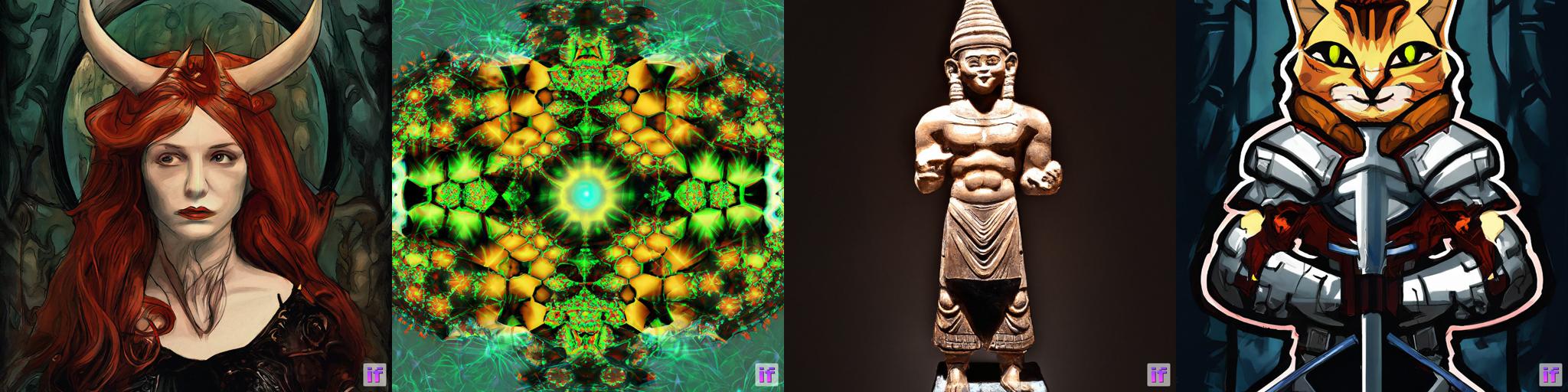}}
  \centerline{IF \cite{gen:IF}}\medskip
\end{minipage}
\begin{minipage}[(b) ]{0.33\linewidth}
  \centering
  \centerline{\includegraphics[width = \textwidth]{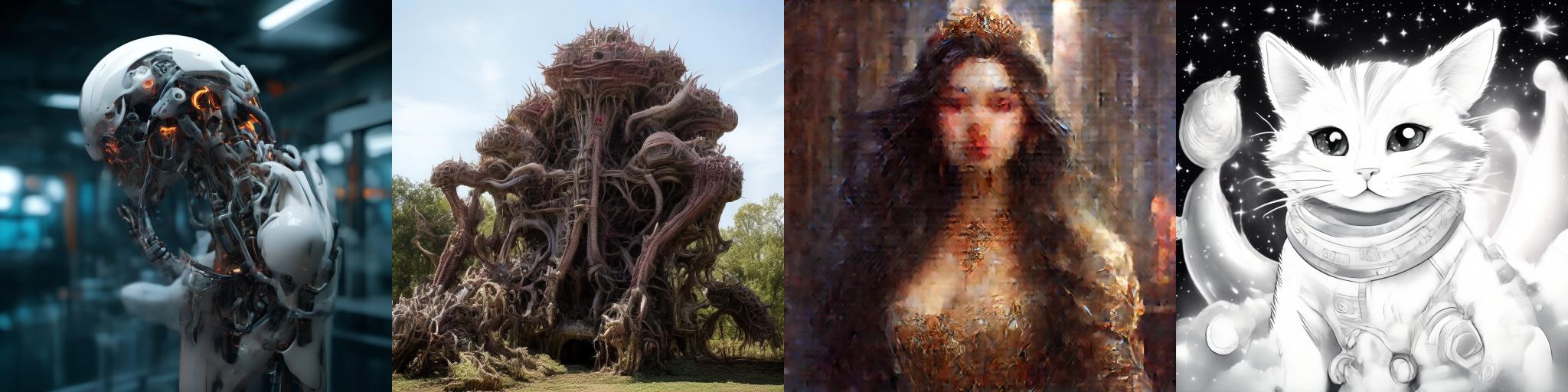}}
  \centerline{LCM Pixart \cite{gen:lcm}}\medskip
\end{minipage}
\begin{minipage}[(b) ]{0.33\linewidth}
  \centering
  \centerline{\includegraphics[width = \textwidth]{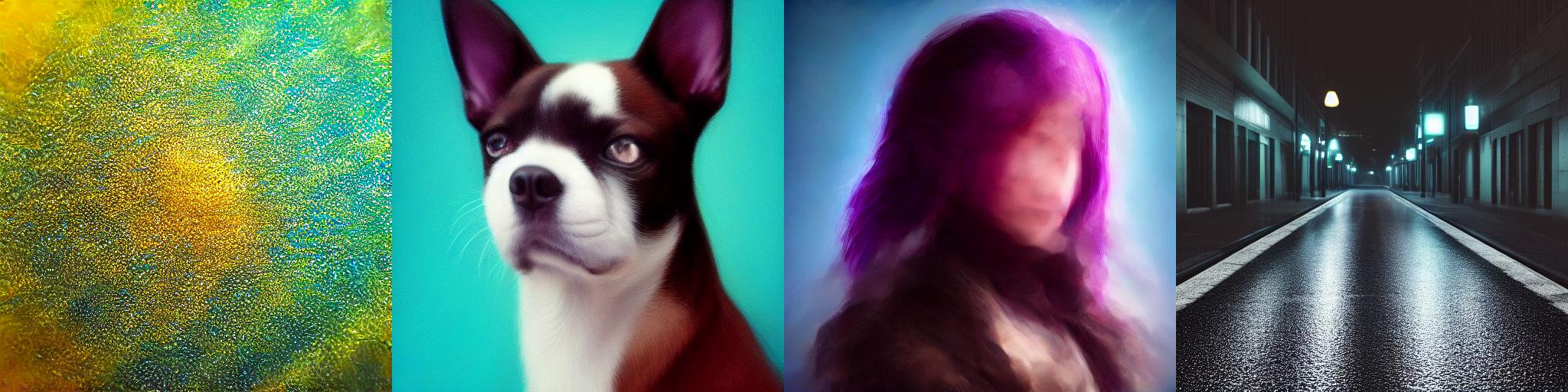}}
  \centerline{LCM SD1.5 \cite{gen:lcm}}\medskip
\end{minipage}
\begin{minipage}[(b) ]{0.33\linewidth}
  \centering
  \centerline{\includegraphics[width = \textwidth]{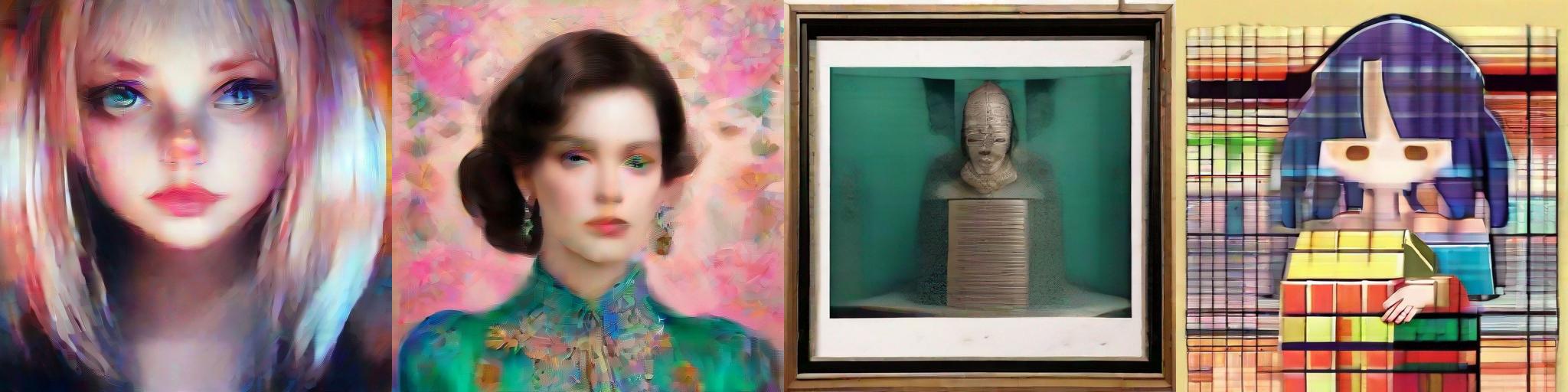}}
  \centerline{LCM SDXL \cite{gen:lcm}}\medskip
\end{minipage}
\begin{minipage}[(b) ]{0.33\linewidth}
  \centering
  \centerline{\includegraphics[width = \textwidth]{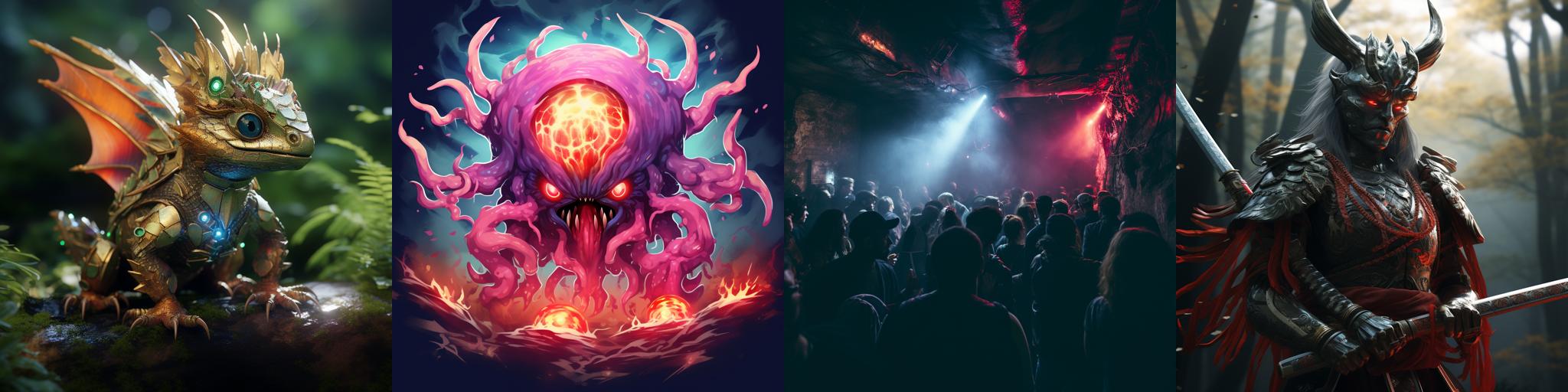}}
  \centerline{Midjourney v5.2 \cite{gen:MJ}}\medskip
\end{minipage}
\begin{minipage}[(b) ]{0.33\linewidth}
  \centering
  \centerline{\includegraphics[width = \textwidth]{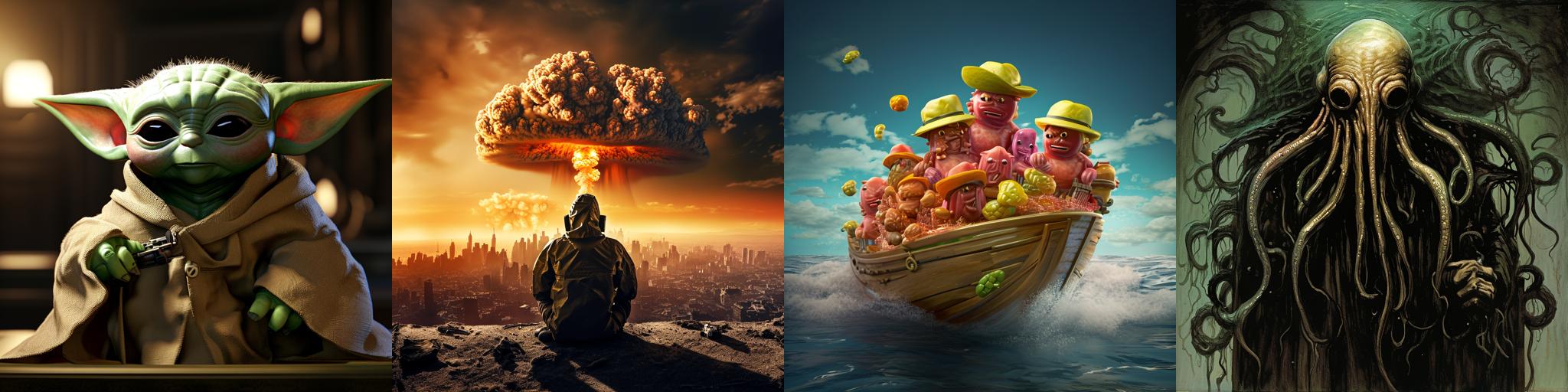}}
  \centerline{Pixart $\alpha$ \cite{gen:pixart}}\medskip
\end{minipage}
\begin{minipage}[(b) ]{0.33\linewidth}
  \centering
  \centerline{\includegraphics[width = \textwidth]{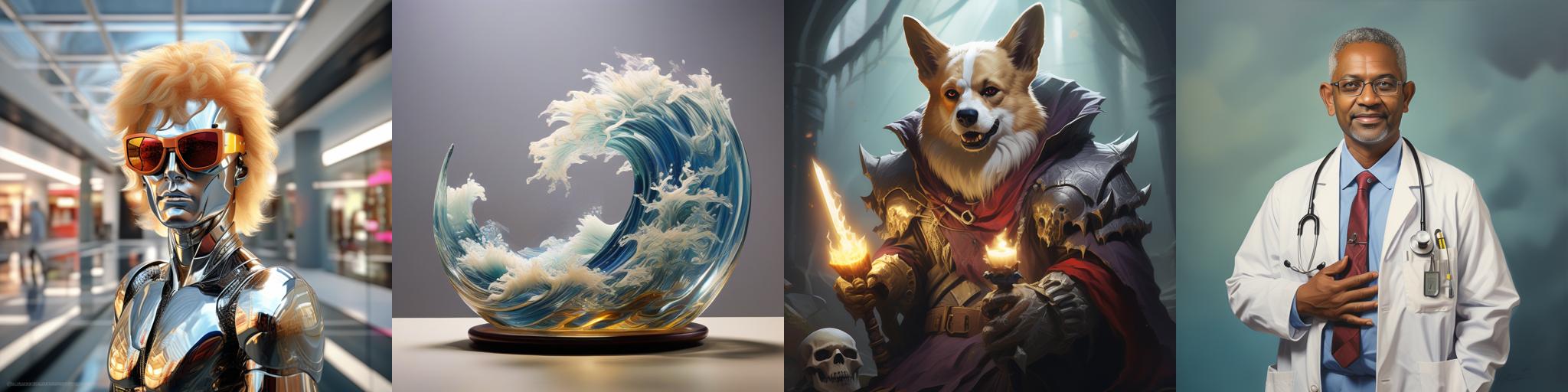}}
  \centerline{Playground v2 \cite{gen:Playground}}\medskip
\end{minipage}
\begin{minipage}[(b) ]{0.33\linewidth}
  \centering
  \centerline{\includegraphics[width = \textwidth]{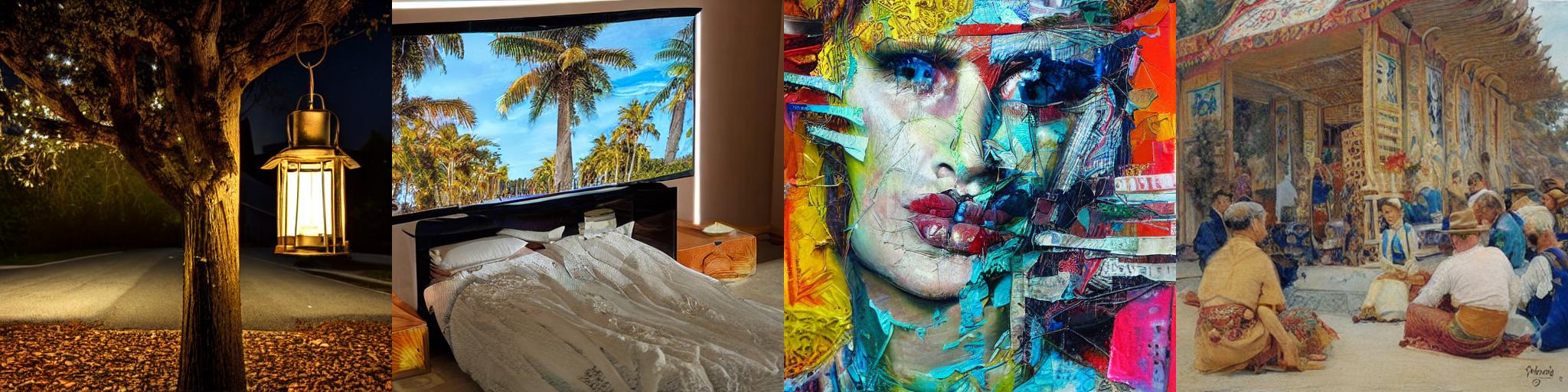}}
  \centerline{SD1.4 \cite{gen:sd}}\medskip
\end{minipage}
\begin{minipage}[(b) ]{0.33\linewidth}
  \centering
  \centerline{\includegraphics[width = \textwidth]{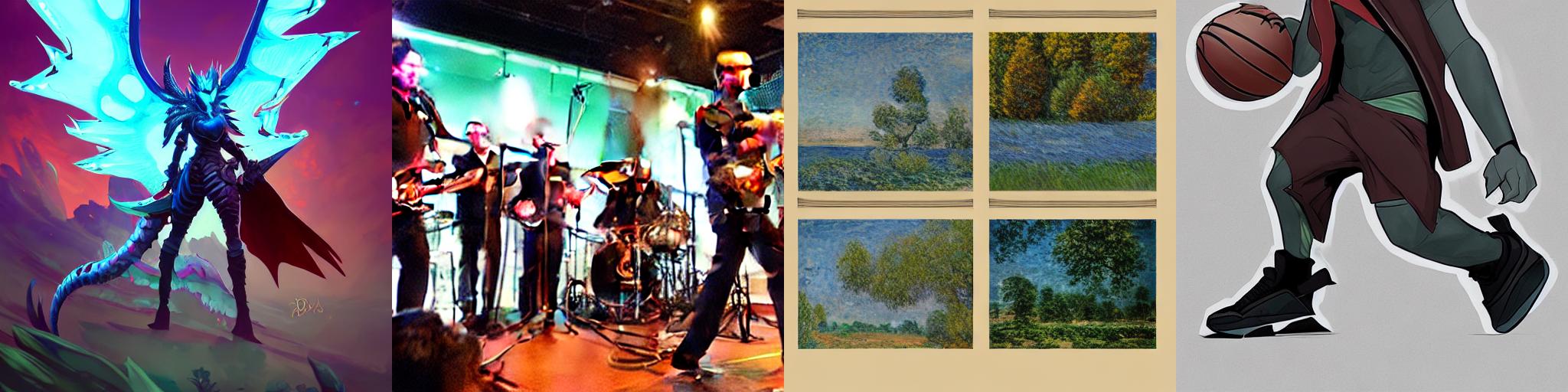}}
  \centerline{SD1.5 \cite{gen:sd}}\medskip
\end{minipage}
\begin{minipage}[(b) ]{0.33\linewidth}
  \centering
  \centerline{\includegraphics[width = \textwidth]{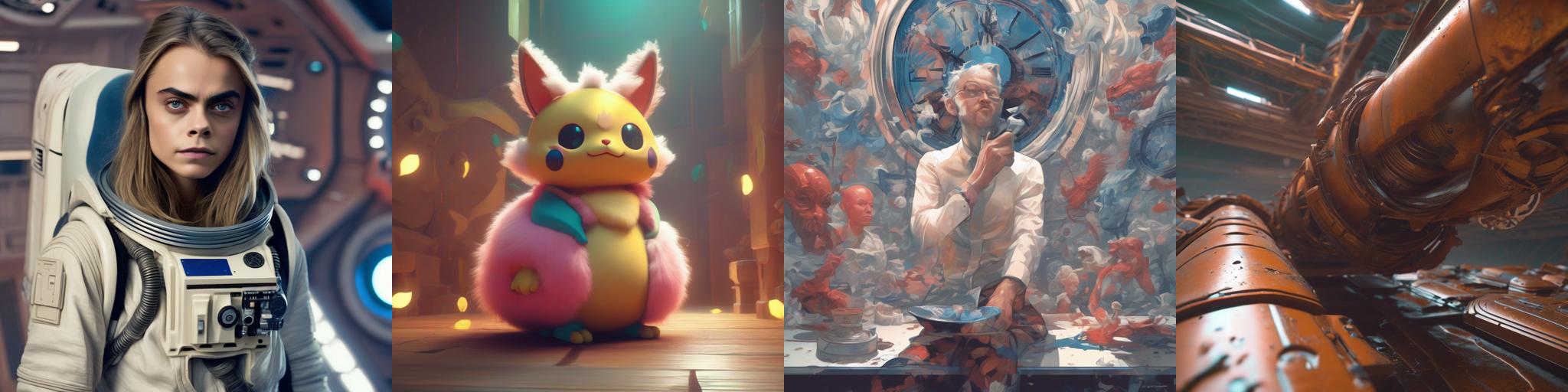}}
  \centerline{SDXL \cite{gen:xl}}\medskip
\end{minipage}
\begin{minipage}[(b) ]{0.33\linewidth}
  \centering
  \centerline{\includegraphics[width = \textwidth]{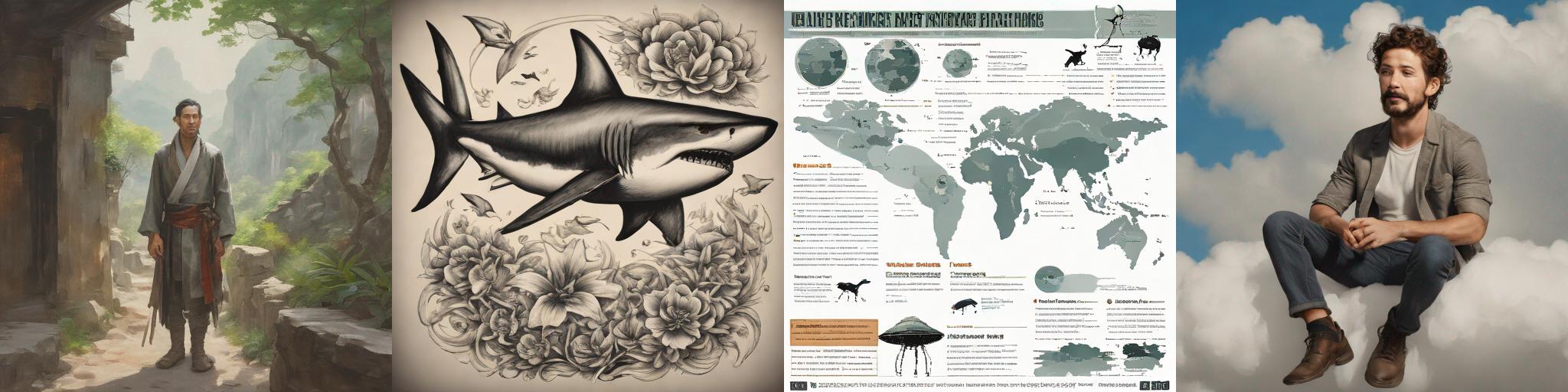}}
  \centerline{SDXL Turbo \cite{gen:turbo}}\medskip
\end{minipage}
\begin{minipage}[(b) ]{0.33\linewidth}
  \centering
  \centerline{\includegraphics[width = \textwidth]{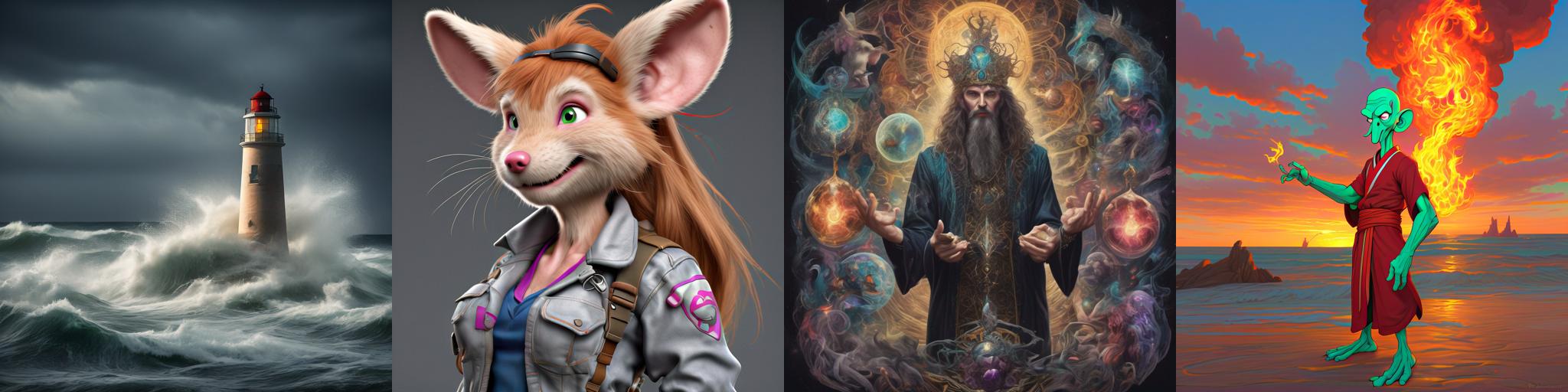}}
  \centerline{SSD1B \cite{gen:ssd-1b}}\medskip
\end{minipage}

\caption{Visulization result of 15 Text-to-Image models in AIGIQA-20K.}
  \label{fig:example}
\end{figure*}

\section{Database Construction}
\label{sec:construction}
\subsection{Hyper-parameter Configuration}
\label{sec:parameter}
The quality of model generation is closely related to the hyper-parameters. Due to the limited computational resources and different settings, these hyper-parameter configurations change frequently in the actual generation process. Among them, insufficient iterations will reduce image detail; too high/low Classifier Free Guidance (CFG) will affect the tradeoff between perceptual/alignment quality; non-square resolution will cause a sharp drop in overall quality. Therefore, before the generation process for each T2I model, our AIGIQA-20K database dynamically set these quality-aware configurations with the following criteria:

\begin{itemize}
    \item Iterations: 50\% as default full epochs, 25\% as $0.5\times$ epochs, and 25\% as $0.25\times$ epochs.
    \item CFG: 50\% as default CFG number, 20\% as $0.5\times$ default CFG, 20\% as $2\times$ default CFG, and 10\% applies $0.5 \sim 2$ default CFG randomly.
    \item Resolution: 50\% as $1:1$ square, four 10\% as $3:4$, $4:3$, $9:16$, $16:9$, and final 10\% as ${9:16} \sim {16:9}$ randomly. The longest edge is set as 512 or 1,024 according to the maximum resolution of the model.
\end{itemize}
where the configuration adjustments for each model are described in the next section.

\subsection{Generative Model Collection}
\label{sec:model}

\begin{figure*}[t]
\begin{minipage}[(b) ]{0.5\linewidth}
  \centering
  \centerline{\includegraphics[width = \textwidth]{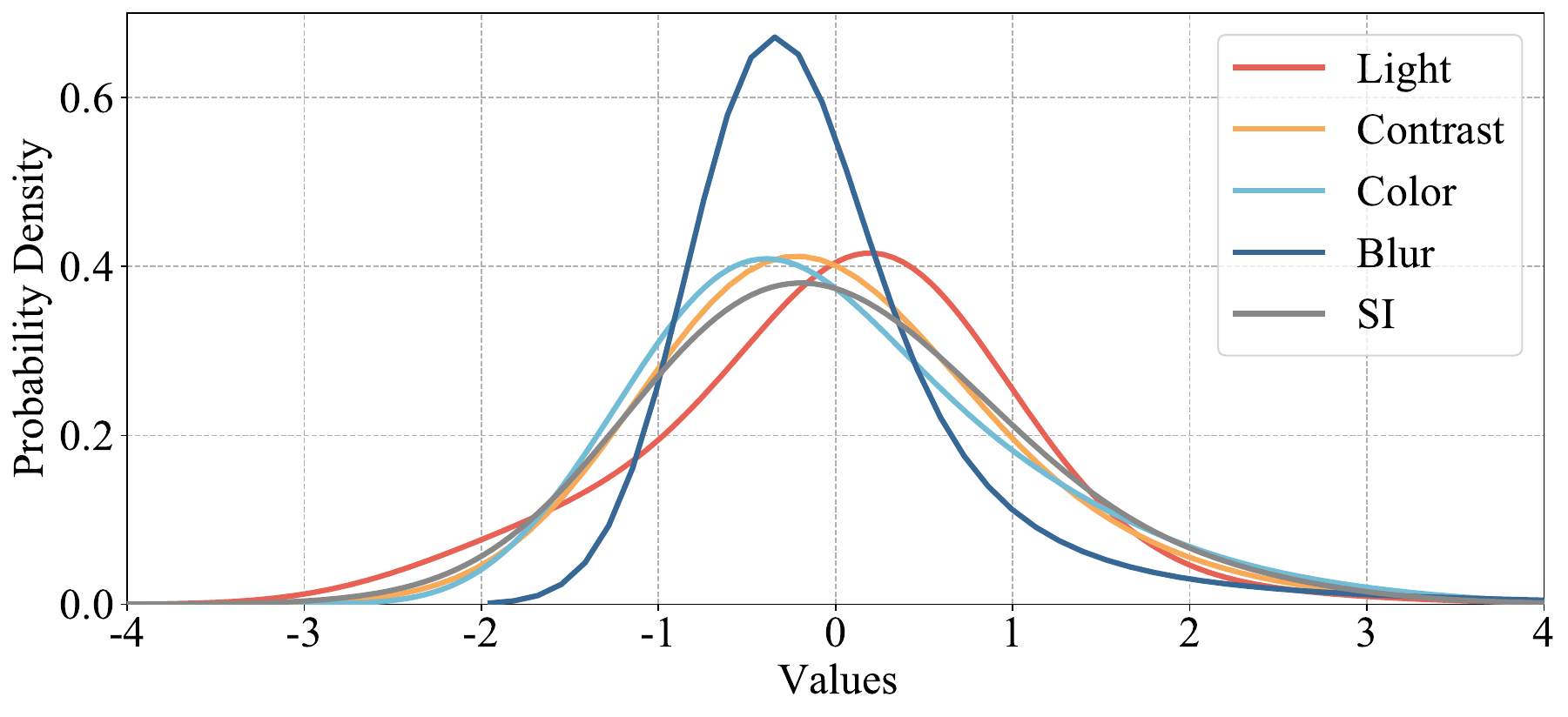}}
  \centerline{(a) All images}\medskip
  \end{minipage}
\begin{minipage}[(b) ]{0.5\linewidth}
  \centering
  \centerline{\includegraphics[width = \textwidth]{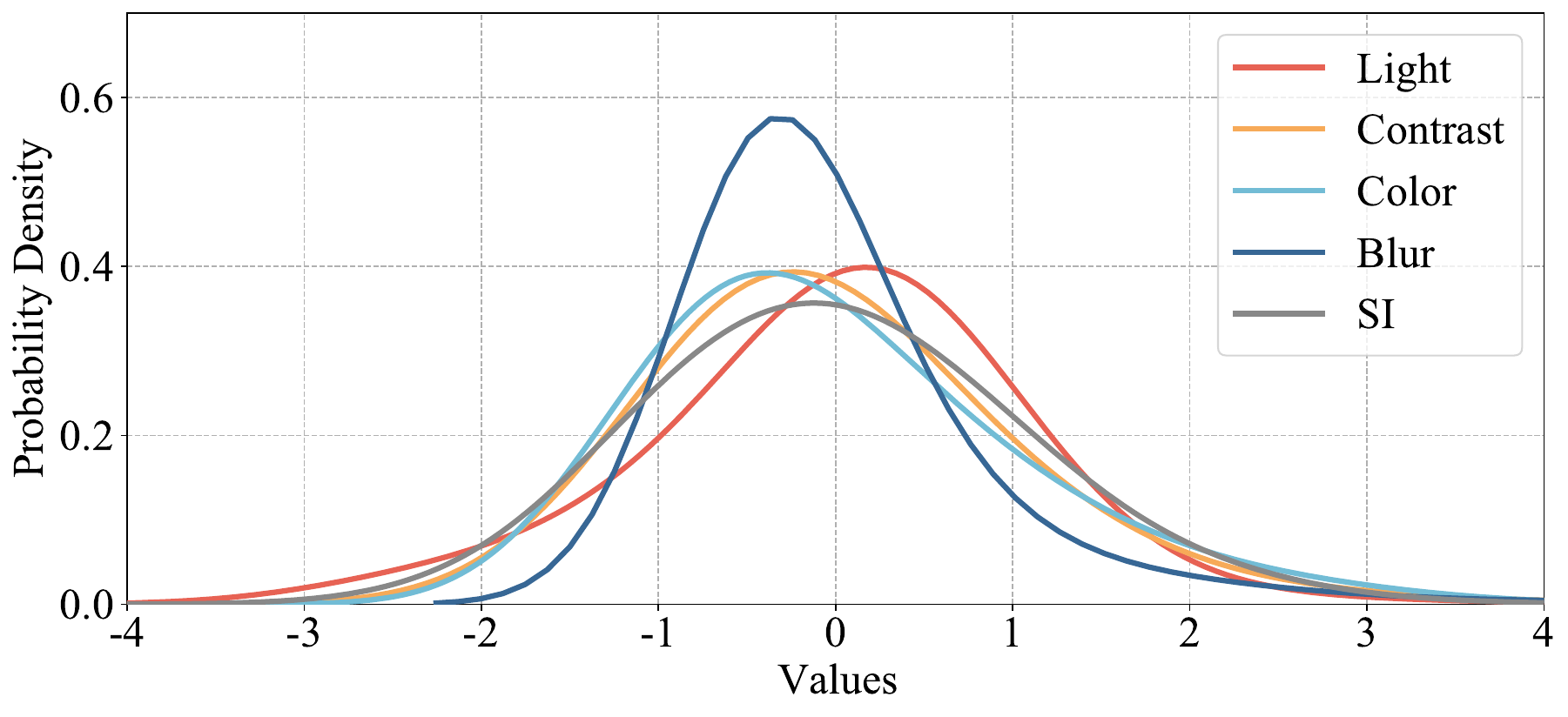}}
  \centerline{(b) Images with wrong CFG}\medskip
\end{minipage}

\begin{minipage}[(b) ]{0.5\linewidth}
  \centering
  \centerline{\includegraphics[width = \textwidth]{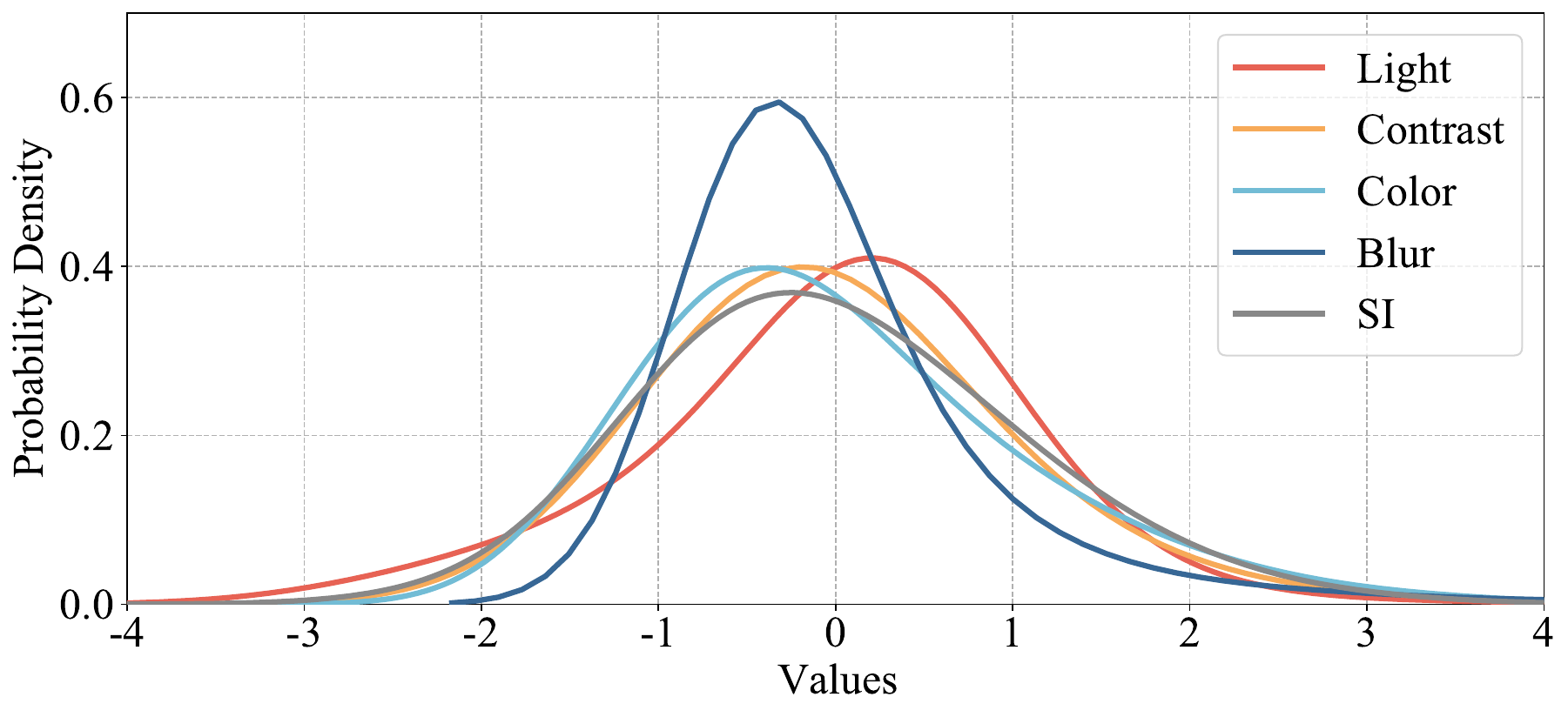}}
  \centerline{(c) Images with limited iterations}\medskip
  \end{minipage}
\begin{minipage}[(b) ]{0.5\linewidth}
  \centering
  \centerline{\includegraphics[width = \textwidth]{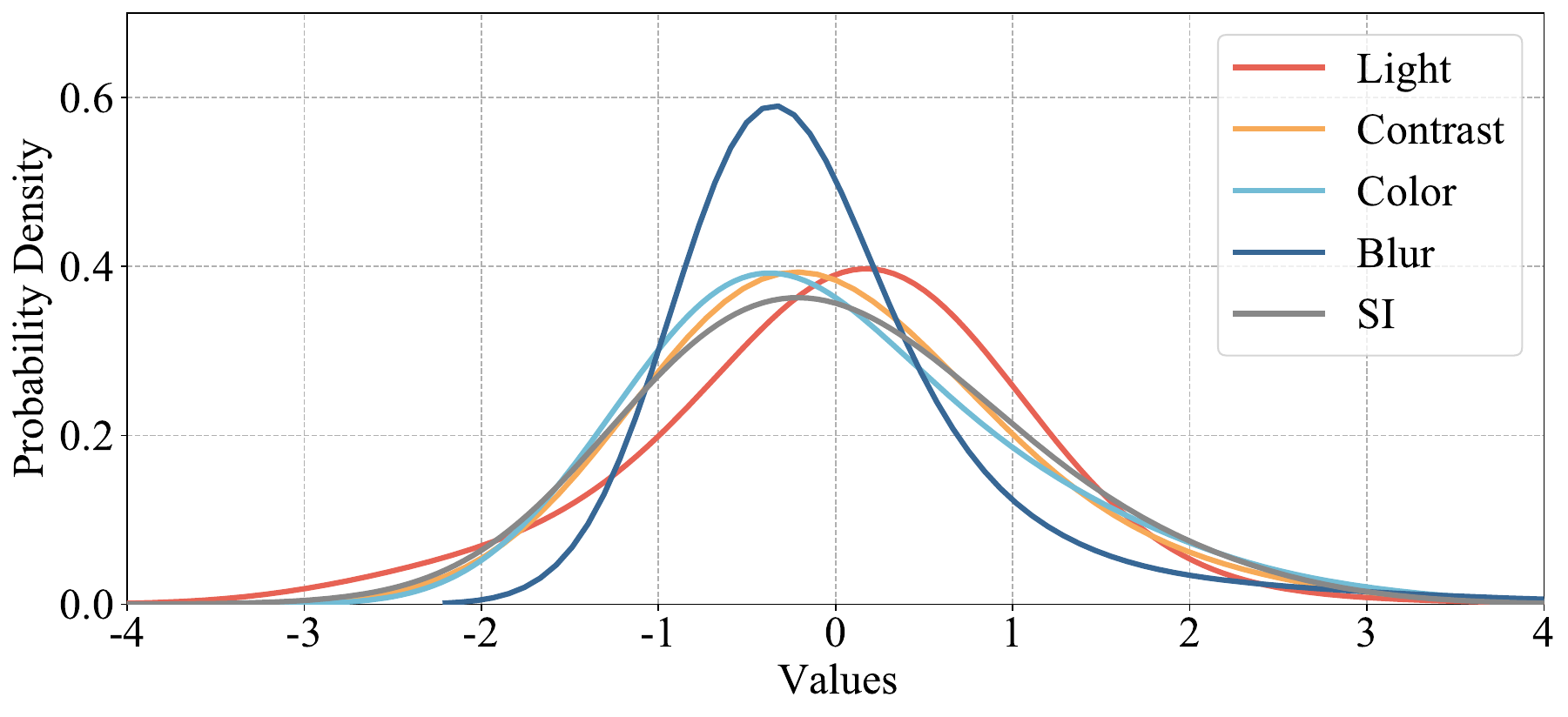}}
  \centerline{(d) Images with non-square resolution}\medskip
\end{minipage}
\caption{Distribution of quality attribute over the AIGIQA-20K database three sub-datasets with abnormal hyper-parameters.
}
\label{fig:feature}
\end{figure*}
Based on the size of previous fine-grained database \cite{ntire:vdpve,ntire:aigcvideo}, the AIGIQA-20K includes 2,000 $\times$ 7+1,000 $\times$ 4+500 $\times$ 4=20,000 images.
Considering that the overall generation effect of the diffusion-based model is well-developed and widely used, we discard previous Generative Adversarial Network (GAN) and Auto-Regressive (AR) \cite{gen:Cogview} models that have been eliminated for the current T2I generation task. To ensure content diversity, our AIGCQA-20K database considered 15 representative T2I generative models in Figure \ref{fig:example}. For each model, with specific configurations in Sec \ref{sec:parameter}, we generate the following number of images:
\begin{itemize}
    \item 2,000 images: Dreamlike, Pixart $\alpha$, Playground v2, SD1.4, SD1.5, SDXL, SSD1B  ~\cite{gen:dream,gen:IF,gen:pixart,gen:Playground,gen:sd,gen:ssd-1b}. These models have strong generalize ability so we change all three hyper-parameters with default iterations as 40.
    \item 1,000 images: LCM Pixart, LCM SD1.5, LCM SDXL, SDXL Turbo ~\cite{gen:lcm,gen:turbo}. At the cost of a fixed CFG, these models use acceleration mechanisms that significantly reduce iteration times. Thus, we only change iterations/resolutions with default iterations as 4.
    \item 500 images: DALLE2, DALLE3, IF, Midjourney v5.2 ~\cite{gen:dalle,gen:IF,gen:MJ}. Adjusting the hyper-parameters drastically reduces the quality of their output. Due to overly complex model structures or closed sources, we set all three parameters to their own default values. 
\end{itemize}
where we generate 20,000 images with different configurations while ensuring 500 results from each model according to the default configuration. Therefore, the database can be used for both IQA tasks and horizontal comparisons of output quality between different models.

\subsection{Prompts Selection}
\label{sec:model}

For AIGI quality databases, prompts are typically from real input or manually designed. Here, AIGIQA-20K relies on the real input of AIGC community users. Firstly, as a large-scale database, AIGIQA-20K has extensively covered inputs of different lengths/themes/styles, eliminating the need to design manual prompts like previous small databases to ensure the diversity of input content. Secondly, using real user input is more in line with the real usage scenarios of AIGC, and the quality score obtained is also more reasonable. Therefore, we selected 30,000 prompt words from DiffusionDB as the original input.
Considering the presence of some junk data in the above prompts, we adopt the following filtering mechanism: (1) Similarity comparison: prompts with 90\% consistent content will be merged; (2) Character detection: Remove consecutive spaces, parentheses, punctuation, and non-UTF-8 encoded characters; (3) NSFW avoidance: First, delete prompts containing sensitive words, and then use GPT-4 \cite{tool:gpt4} to delete NSFW prompts in semantic level. From this, we filtered out 20,000 prompts as inputs for T2I models.

\subsection{Feature Analysis}
\label{sec:feature}

After generating images from prompts and hyper-parameters above, to evaluate the impact of different configurations on the images, we calculated the distribution of five quality-related attributes in Figure \ref{fig:feature} for the entire database, adjusted subsets of iteration, cfg, and resolution. The quality-related attributes include light, contrast, color, blur, and Spatial Information (SI, representing the content diversity of the image). Detailed explanations of these attributes can be found in work \cite{other:define}. As previous works ~\cite{database:agiqa3k,other:misc} state, AIGIs have more extreme blur distribution than NSIs, while the other four attributes are distributed more evenly. In addition, we also found that for each subset of hyper-parameter anomalies, the maximum probability values never exceed 0.6; however, for the entire AIGIQA-20K, its distribution curve is sharper. This difference indicates a significant gap between the attributes of default and abnormal hyper-parameter subsets, which indirectly demonstrates the strong correlation between hyper-parameters and image quality while demonstrating the necessity of such adjustments.

\section{Subjective Experiment}
\label{sec:subjective}
\subsection{Experimental Procedures}
\begin{figure}[t]
\centering
\includegraphics[width=0.4\textwidth]{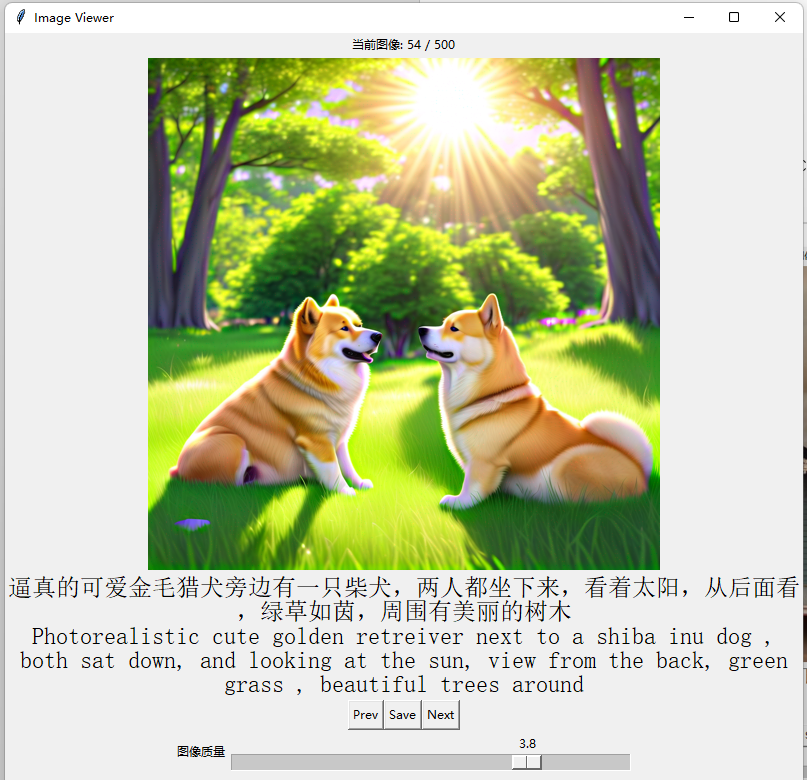}
\caption{User interface for subjective quality assessment. The image is given together with its correlated prompt. The score is an overall consideration of perceptual quality and alignment.}
\label{fig:ui}
\end{figure}
Compliant with the ITU-R BT.500-13 \cite{other:itu} standard, we invited 21 subjects (12 male, 9 female) in this subjective experiment with normal lighting levels. AIGIs are presented on the iMac display together with the prompt in random order on the screen, with a resolution of up to 4096 $\times$ 2304. Both prompt and image are accessible for subjective, with a translation of their mother tongue like Figure \ref{fig:ui}. Considering the average between perceptual quality and T2I alignment, subjects were asked to give an overall score within the range of [0, 5], where each one-point interval stands for poor, bad, fair, good, or excellent quality.

\begin{figure}[t]
\includegraphics[width=0.5\textwidth]{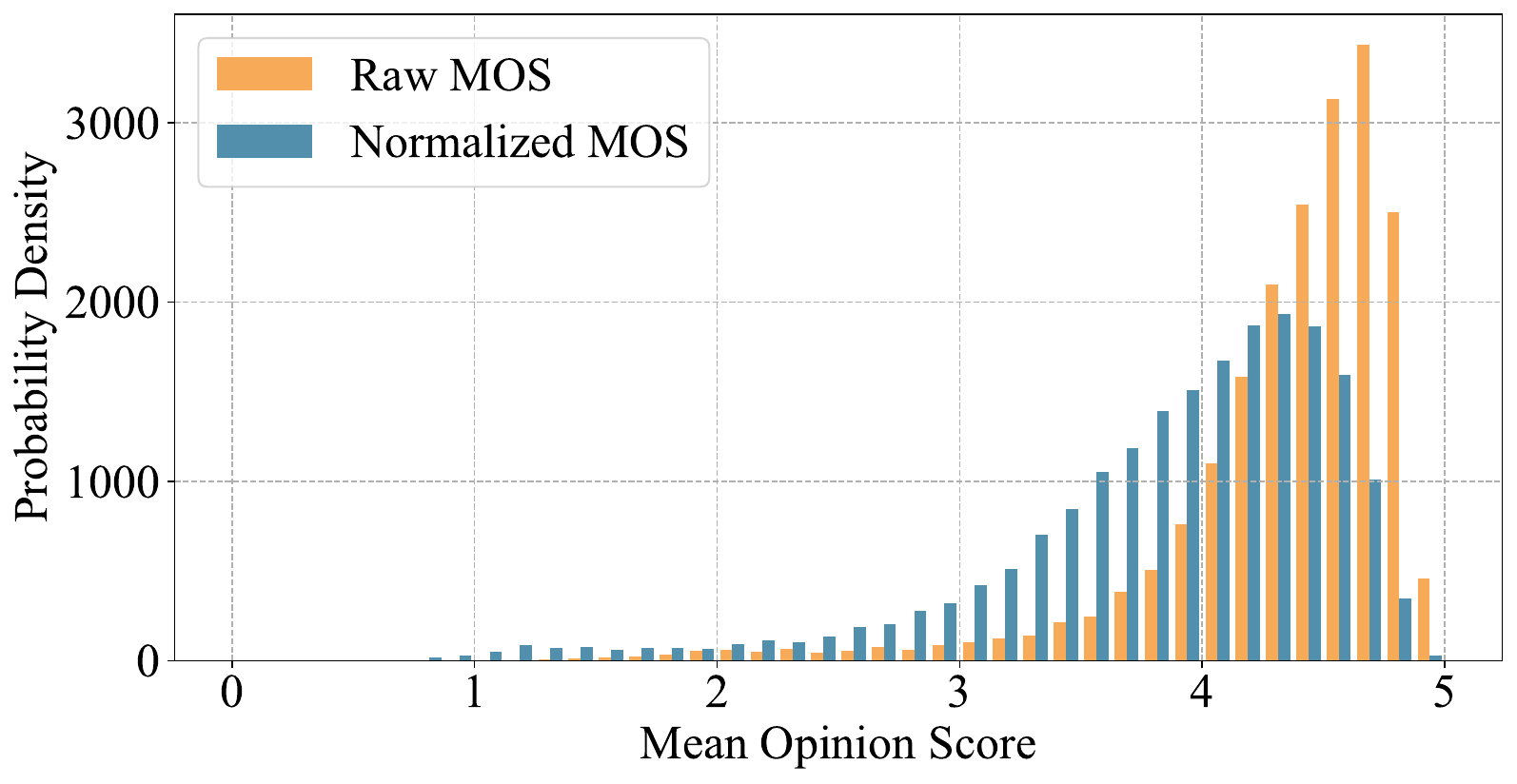}
\caption{Distribution of raw and logarithmic normalized MOSs, where the logarithmic function unifies the entire distribution.}
\label{fig:mos}
\end{figure}

\subsection{Data Processing}
In case of visual fatigue, we split the database into $g \in [0,39]$ groups including $M=500$ images each, while limiting the experiment time to half an hour. After collecting 21$\times$20,000=42,000 quality ratings, we compute the Spearman Rank-order Correlation Coefficient (SRoCC) between them and the global average and remove the outliers with SRoCC lower than 0.6. Then we normalize the average score $s$ for between each session to avoid inter-session scoring differences as:
\begin{equation}
s_{ij}(g) = r_{ij}(g) - \frac{1}{M}\sum_{i=0}^{g\cdot M -1} r_{ij} + 2.5,
\end{equation}
where $(i,j)$ represent the index of the image and viewer and $r$ stands for raw score. Then subjective scores are converted to Z-scores $z_{ij}$ by:
\begin{equation}
z_{ij}=\frac{s_{ij}-\mu_j}{\sigma_j},
\end{equation}
where $\mu_j=\frac{1}{N}\sum_{i=0}^{N-1} s_{ij}$, $\sigma_j=\sqrt{\frac{1}{N-1}\sum_{i=0}^{N-1}(s_{ij}-\mu_i)^2}$ and $N=40$ is the number of subjects. Finally, the MOS of image $j$ is computed with the following formula:
\begin{equation}
\left\{ \begin{array}{l}
MOS_i={\rm log}(\frac{1}{N}\sum_{j=0}^{N-1} (z_{ij})+1) \\
MOS=5 \cdot {\rm norm}(MOS),
\end{array} \right.
\end{equation}
where ${\rm norm}(\cdot)$ indicates 0-1 normalization is a traditional data-processing technique, but logarithmic function ${\rm log}(\cdot)$ is specially designed for AIGI. As shown in Figure \ref{fig:mos}, the raw MOS data shows severe right deviation, almost all of which are concentrated in the 4 to 5 interval. However, after logarithmic processing, the distribution of MOS becomes more uniform. This highly differentiated score is more suitable for IQA tasks.

\begin{figure}[tbph]
\begin{minipage}[(b) ]{\linewidth}
  \centering
  \centerline{\includegraphics[width = \textwidth]{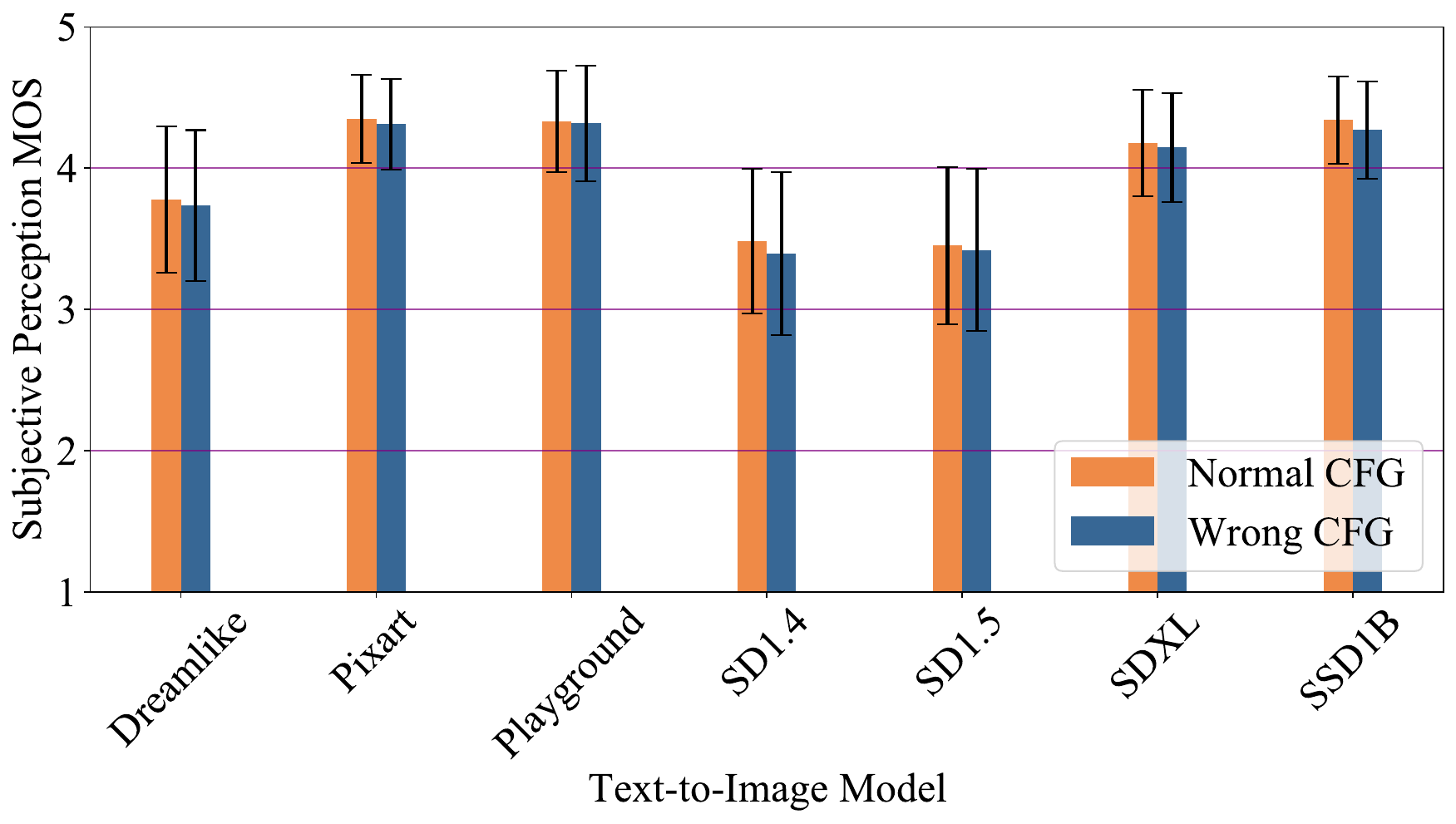}}
  \centerline{(a) Images with default and wrong CFG}\medskip
  \end{minipage}
\begin{minipage}[(b) ]{\linewidth}
  \centering
  \centerline{\includegraphics[width = \textwidth]{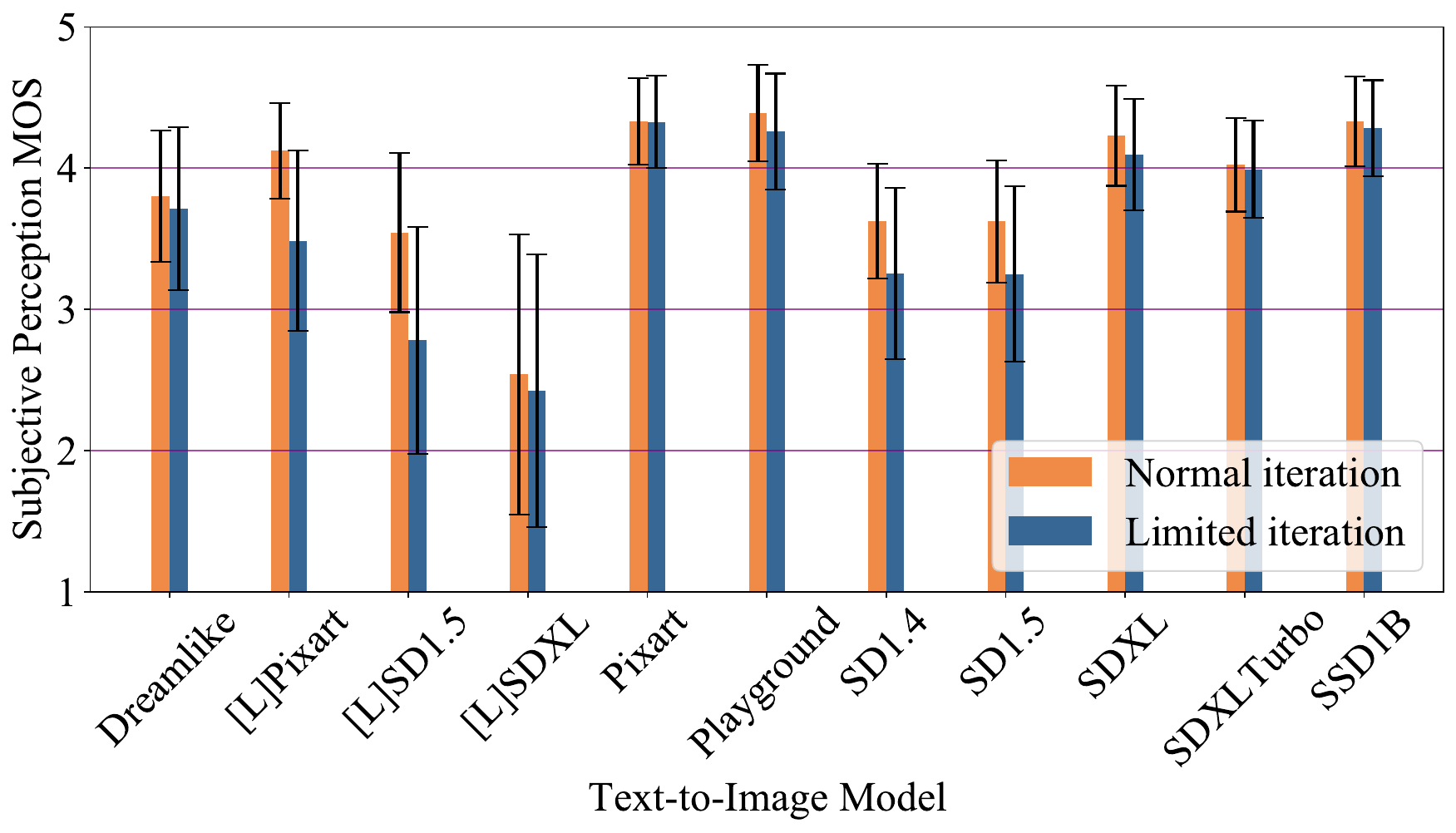}}
  \centerline{(b) Images with default and limited iterations}\medskip
\end{minipage}
\begin{minipage}[(b) ]{\linewidth}
  \centering
  \centerline{\includegraphics[width = \textwidth]{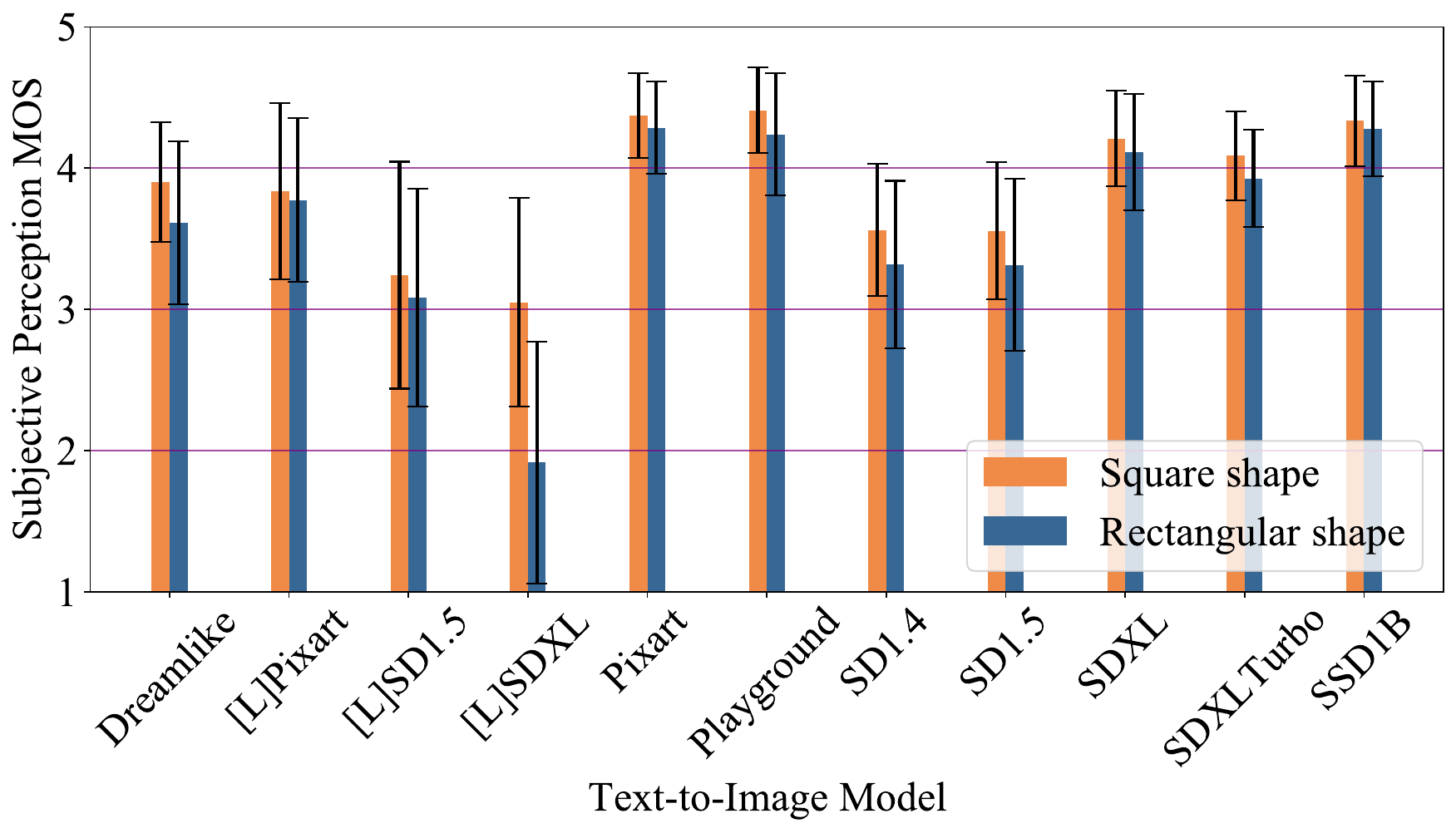}}
  \centerline{(c) Images with default (square) and rectangular resolutions}\medskip
  \end{minipage}
\caption{Subjective quality score of images with default and abnormal hyper-parameters. For all T2I models with abnormal hyper-parameters, the subjective quality decreases to a certain extent compared to default. (\{L\} for LCM)
}
\label{fig:anal-para}
\vspace{-1mm}
\end{figure}
\begin{figure*}[t]
\includegraphics[width=\textwidth]{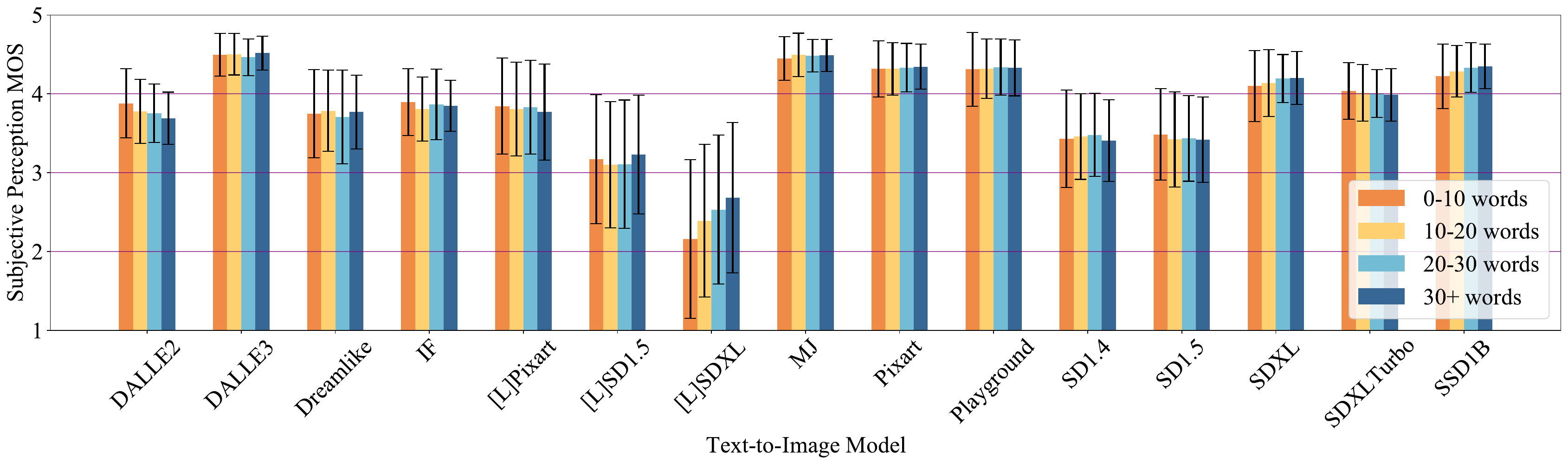}
\caption{Subjective quality score of images with different T2I models and prompt length. (\{L\} for LCM)}
\label{fig:anal-word}
\end{figure*}
\subsection{Data Analysis}
With the explosion of T2I models, their generative quality has become an unresolved issue. Compared to objective indicators, subjective evaluation indicators can better reflect real human preferences. Based on the large-scale and fine-grained subjective quality ratings in the AIGIQA-20K database, we conducted an in-depth analysis of this issue and summarized the influencing factors of AIGI subjective quality as follows:
\begin{itemize}
    \item T2I model: Generative models themselves are the primary determinant of AIGI quality. Under the same input prompt, the generation quality of different models varies greatly. The lower limit of the latest models even outperforms the upper limit of old models.
    \item Prompt: The prompt has a certain impact on the quality of AIGI. 
%    Previous work has shown that training data determines the quality of T2I model generation for a certain style. In addition,
    Different models apply their own text encoders, some are good at generating short prompts, while others are suitable for long prompts.
    \item Hyper-parameters: The internal parameters of a model can profoundly affect the quality of AIGI. As Sec \ref{sec:parameter} listed, CFG, number of iterations, and resolution can all cause AIGI quality fluctuations.
\end{itemize}
For the influence of hyperparameters, we divided the entire AIGIQA-20K database into two parts based on CFG, iteration, and resolution, and compared their subjective quality distribution under normal (default value) and abnormal (configuration in Sec \ref{sec:parameter}) states, as shown in Figure \ref{fig:anal-para}. 
Firstly, for CFG, there has been a slight decrease in quality after adjustment compared to the default value. This is because CFG reflects the trade-off between perceived quality and fit. The larger the value, the more the model values alignment, and vice versa, the more emphasis is placed on perception. However, regardless of which side reduces and which side increases, the quality score obtained by considering both factors will inevitably decrease. This demonstrates the validity of the default CFG values in each T2I model and it is not recommended to adjust them arbitrarily.
Secondly, for the number of iterations, the quality with limited iterations has decreased to varying degrees against full iterations. This indicates when iterations are insufficient, the AIGIs may lack certain details, leading to a decrease in quality. Compared to various models, the most advanced Pixart, Playground, SDXL, SDXL Turbo, and SSD1B ~\cite{gen:pixart,gen:Playground,gen:xl,gen:turbo,gen:ssd-1b} have the strongest robustness to this descent. For models with the LCM acceleration mechanism, the number of iterations is already as low as 4, and further reducing iterations will cause significant quality damage.
Thirdly, for shapes, the quality of generating irregular shapes is also lower than that of squares. Since the target outputs during model training are all squares, it is expected that generating rectangular images is not ideal. Horizontally comparing others, LCM Pixart, Pixart, and SSD1B ~\cite{gen:lcm,gen:pixart,gen:ssd-1b} have the strongest robustness to such descent. Overall, the newer the model, the better its support for non-square outputs.

For the T2I models themselves and prompts, Figure \ref{fig:anal-word} lists the subjective quality of all 15 models at different prompt lengths. All hyper-parameters are set by default for a fair comparison. Firstly, by comparing the various models at the level of human preferences, the most advanced models currently available are DALLE3, Midjourney, Pixart, and Playground ~\cite{gen:dalle,gen:MJ,gen:pixart,gen:Playground};
%In addition, both SDXL and SDXL Turbo have achieved an average quality score of 4 or above;
and all other models have certain quality defects. For the acceleration mechanism, SDXL Turbo \cite{gen:turbo} is the most successful, as it reduces the iterations by 10 times at a quality cost below 0.2; In contrast, after 10 times acceleration, the output quality of the LCM \cite{gen:lcm} model is far inferior to the original, especially the acceleration effect on SDXL is extremely poor. Secondly, except for LCM SDXL, the T2I model has slightly better quality in generating short prompts than long texts. This comes from the limitation of the number of embedding tokens in the model text encoder. For example, CLIP only supports 77 tokens in absolute terms; Even worse, existing research \cite{other:longclip} indicates that it gradually fails even from the 20th token onwards. Therefore, the defect in alignment resulted in a decrease in overall scores. In summary, to improve the output quality of the T2I model. Users should set appropriate hyper-parameters while developers need to design more powerful models and enhance their support for long text encoding and multiple resolutions.

\section{Experiment}
\label{sec:experiment}
\subsection{Experiment Settings}
We first randomly split the AIGIQA-20K into training/validation/test sets according to the ratio of 7:1:2, with 14,000/2,000/4,000 AIGIs respectively. To benchmark the performance of quality metrics, three global indicators, including SRoCC, Kendall Rank-order Correlation Coefficient (KRoCC), and Pearson Linear Correlation Coefficient (PLCC) are applied to evaluate the consistency between the objective quality score and the subjective MOS, among which the SRoCC and KRoCC represent the prediction monotonicity while the PLCC measures the prediction accuracy. To map the objective predicted scores to subjective MOSs, a standard five-parameter logistic function is applied as follows:
\begin{equation}
\hat{X}=\alpha_{1}\left(0.5-\frac{1}{1+e^{\alpha_{2}\left(X-\alpha_{3}\right)}}\right)+\alpha_{4} X+\alpha_{5},
\end{equation}
where $\alpha_{1 \sim 5}$ represent the parameters for fitting, $X$ and $\hat{X}$ stand for predicted and fitted scores respectively.

\subsection{Benchmark Models}
We apply 16 mainstream AIGI quality benchmarks for comparison, including both perception and alignment metrics. For perception, 12 IQA metrics are selected in the experiment. Including brisque \cite{quality:brisque}, clipiqa \cite{quality:CLIPIQA}, cnniqa \cite{quality:CNNIQA}, dbcnn \cite{quality:DBCNN}, hypetiqa \cite{quality:HyperIQA}, liqe \cite{quality:LIQE}, musiq \cite{quality:musiq}, niqe \cite{quality:NIQE}, qalign \cite{quality:q-align}, topiq \cite{quality:topiq}, unique \cite{quality:unique}, and wadiqam \cite{quality:wadiqam}. For the lower-better indexes (brisque, niqe), their score is reversed. These indicators mainly focus on the image itself, in the absence of the absence of text prompts. For alignment, we take 4 advanced metrics for AIGI quality. The clip \cite{align:clip} mainly considers T2I alignment between AIGIs and prompts while the hps \cite{database/align:HPS}, imagereward \cite{database/align:ImageReward}, and picscore \cite{database/align:PickAPic} also take the perceptual quality as an auxiliary indicator. Most of them are validated as zero-shot models while cnniqa, clipiqa, and dbcnn ~\cite{quality:CNNIQA,quality:CLIPIQA,quality:DBCNN} are trained/validated on the target set (repeating 10 times with the average result as final performance), and all the experimental results are from the testing test. The Adam optimizer \cite{other:adam} (with an initial learning rate of 0.00001 and batch size 128) is used for 100-epochs finetune training on an NVIDIA RTX A6000 GPU.

\subsection{Performance Discussion}
Table \ref{tab:train} shows the performance of different IQA methods across the entire AIGIQA-20K database. The most powerful quality metric currently based on multimodal large language models, q-align, is used as the baseline. We use three classic quality metrics based on deep learning, finetuning them on the training set before testing, and comparing them with the baseline. Experimental data shows that fine-tuning is of great significance for AIGC quality assessment. The performance of the three methods after training has significantly improved, with SRoCC and PLCC improving by 0.4 and KRoCC at 0.3. The performance of clipiqa and dbcnn after training has exceeded the baseline. This is because most IQA models are designed for NSIs, and when they are migrated to AIGIs, it is necessary to update the internal parameters of the model to ensure good performance. Although the fine-tuning effect is good, for promoting the application of IQA in the whole AIGC community, this indicator cannot be completely dependent on fine-tuning training. A powerful zero-shot model like qalign needs further development in the future.

\begin{table}[t]
\centering
\caption{Performance results on the AIGIQA-20K database using zero-shot or finetuned metrics. The zero-shot qalign \cite{quality:q-align} is set as the baseline. [Key: {\bf\textcolor{red}{Best}, \bf\textcolor{blue}{Second Best}}]}
\label{tab:train}
\begin{tabular}{l|rrr}
\toprule
\multicolumn{1}{c}{\textbf{Metric}}  & \multicolumn{1}{c}{SRoCC} & \multicolumn{1}{c}{KRoCC} & \multicolumn{1}{c}{PLCC} \\ \midrule
qalign \cite{quality:q-align}          & 0.7461                    & 0.5511                    & \bf\textcolor{blue}{0.7416}                   \\
clipiqa \cite{quality:CLIPIQA}          & 0.3311                    & 0.2257                    & 0.4829                   \\
clipiqa+finetune & \bf\textcolor{blue}{0.7863}                    & \bf\textcolor{blue}{0.5828}                    & 0.7117                   \\
cnniqa \cite{quality:CNNIQA}          & 0.3299                    & 0.2244                    & 0.3666                   \\
cnniqa+finetune  & 0.5968                    & 0.4183                    & 0.5913                   \\
dbcnn \cite{quality:DBCNN}           & 0.4710                    & 0.3244                    & 0.5120                   \\
dbcnn+finetune   & \bf\textcolor{red}{0.8506}                    & \bf\textcolor{red}{0.6617}                    & \bf\textcolor{red}{0.8688}   \\ \bottomrule               
\end{tabular}
\end{table}\vspace{1mm}

\begin{table*}[tbph]

\caption{Performance results on different AIGIQA-20K sub-database using zero-shot perceptual quality or alignment metrics. The data is split by default/abnormal CFG, iteration, and resolution. [Key: {\bf\textcolor{red}{Best}, \bf\textcolor{blue}{Second Best}}]}
\label{tab:main}
\centering
\begin{tabular}{ll|rrr|rrr|rrr}
\toprule
\multicolumn{2}{c|}{\textbf{Group}}                                                    & \multicolumn{3}{c|}{Default CFG}                                                   & \multicolumn{3}{c|}{Default iteration}                                             & \multicolumn{3}{c}{Default resolution}                                                 \\ \hline
\multicolumn{1}{c|}{\textbf{Type}}              & \multicolumn{1}{c|}{\textbf{Metric}} & \multicolumn{1}{c}{SRoCC} & \multicolumn{1}{c}{KRoCC} & \multicolumn{1}{c|}{PLCC} & \multicolumn{1}{c}{SRoCC} & \multicolumn{1}{c}{KRoCC} & \multicolumn{1}{c|}{PLCC} & \multicolumn{1}{c}{SRoCC} & \multicolumn{1}{c}{KRoCC} & \multicolumn{1}{c}{PLCC} \\ \hline
\multicolumn{1}{l|}{\multirow{12}{*}{\begin{tabular}[c]{@{}l@{}}Perce\\-ption\end{tabular}}}  & brisque \cite{quality:brisque}                              & 0.2755                    & 0.1874                    & 0.2933                    & 0.2259                    & 0.1526                    & 0.2772                    & 0.1883                    & 0.1261                    & 0.1900                   \\
\multicolumn{1}{l|}{}                           & clipiqa \cite{quality:CLIPIQA}                              & 0.3889                    & 0.2677                    & 0.5375                    & 0.2522                    & 0.1719                    & 0.4312                    & 0.2964                    & 0.2015                    & 0.4041                   \\
\multicolumn{1}{l|}{}                           & cnniqa \cite{quality:CNNIQA}                              & 0.3289                    & 0.2238                    & 0.3691                    & 0.3350                    & 0.2274                    & 0.3605                    & 0.3102                    & 0.2088                    & 0.2889                   \\
\multicolumn{1}{l|}{}                           & dbcnn \cite{quality:DBCNN}                               & 0.5051                    & 0.3489                    & 0.5304                    & 0.4492                    & 0.3085                    & 0.4673                    & 0.4380                    & 0.2976                    & 0.4307                   \\
\multicolumn{1}{l|}{}                           & hyperiqa \cite{quality:HyperIQA}                            & 0.4390                    & 0.2990                    & 0.4928                    & 0.3528                    & 0.2393                    & 0.4073                    & 0.3730                    & 0.2526                    & 0.4096                   \\
\multicolumn{1}{l|}{}                           & liqe \cite{quality:LIQE}                                & 0.4925                    & 0.3391                    & 0.5554                    & 0.3675                    & 0.2513                    & 0.4445                    & 0.4030                    & 0.2746                    & 0.4441                   \\
\multicolumn{1}{l|}{}                           & musiq \cite{quality:musiq}                               & 0.5111                    & 0.3546                    & 0.5848                    & 0.3833                    & 0.2616                    & 0.4418                    & 0.4146                    & 0.2837                    & 0.4889                   \\
\multicolumn{1}{l|}{}                           & niqe \cite{quality:NIQE}                                & 0.1900                    & 0.1266                    & 0.3120                    & 0.1999                    & 0.1348                    & 0.2977                    & 0.0769                    & 0.0516                    & 0.1737                   \\
\multicolumn{1}{l|}{}                           & qalign \cite{quality:q-align}                               & \bf\textcolor{red}{0.7721}                    & \bf\textcolor{red}{0.5764}                    & \bf\textcolor{red}{0.7629}                    & \bf\textcolor{red}{0.7145}                    & \bf\textcolor{red}{0.5206}                    & \bf\textcolor{red}{0.6813}                    & \bf\textcolor{red}{0.7333}                    & \bf\textcolor{red}{0.5383}                    & \bf\textcolor{red}{0.7178}                   \\
\multicolumn{1}{l|}{}                           & topiq \cite{quality:topiq}                               & 0.5064                    & 0.3491                    & 0.5292                    & 0.4374                    & 0.2998                    & 0.4487                    & 0.4706                    & 0.3228                    & 0.4663                   \\
\multicolumn{1}{l|}{}                           & unique \cite{quality:unique}                              & 0.3038                    & 0.2041                    & 0.3843                    & 0.1595                    & 0.1075                    & 0.2127                    & 0.2245                    & 0.1510                    & 0.2974                   \\
\multicolumn{1}{l|}{}                           & wadiqam \cite{quality:wadiqam}                             & 0.2821                    & 0.1905                    & 0.2855                    & 0.2847                    & 0.1916                    & 0.2907                    & 0.2516                    & 0.1690                    & 0.2351                   \\ \hline
\multicolumn{1}{l|}{\multirow{4}{*}{\begin{tabular}[c]{@{}l@{}}Align\\-ment\end{tabular}}} & clip \cite{align:clip}                                & 0.4701                    & 0.3656                    & 0.5341                    & 0.3804                    & 0.2969                    & 0.4733                    & 0.3309                    & 0.2580                    & 0.2673                   \\
\multicolumn{1}{l|}{}                           & hps \cite{database/align:HPS}                                 & 0.6749                    & 0.4899                    & 0.6111                    & 0.6288                    & 0.4514                    & 0.5052                    & 0.6214                    & 0.4434                    & 0.4865                   \\
\multicolumn{1}{l|}{}                           & imagereward \cite{database/align:ImageReward}                         & 0.6597                    & 0.4767                    & 0.7162                    & 0.6150                    & 0.4387                    & \bf\textcolor{blue}{0.6691}                    & 0.6098                    & 0.4316                    & \bf\textcolor{blue}{0.6579}                   \\
\multicolumn{1}{l|}{}                           & picscore \cite{database/align:PickAPic}                                 & \bf\textcolor{blue}{0.7009}                    & \bf\textcolor{blue}{0.5093}                    & \bf\textcolor{blue}{0.7201}                    & \bf\textcolor{blue}{0.6474}                    & \bf\textcolor{blue}{0.4642}                    & 0.6638                    & \bf\textcolor{blue}{0.6458}                    & \bf\textcolor{blue}{0.4617}                    & 0.6474                  \\ \bottomrule
\end{tabular}

\vspace{4mm}

\begin{tabular}{ll|rrr|rrr|rrr}
\toprule
\multicolumn{2}{c|}{\textbf{Group}}                                                    & \multicolumn{3}{c|}{Wrong CFG}                                                   & \multicolumn{3}{c|}{Limited iteration}                                             & \multicolumn{3}{c}{Rectangular resolution}                                                 \\ \hline
\multicolumn{1}{c|}{\textbf{Type}}              & \multicolumn{1}{c|}{\textbf{Metric}} & \multicolumn{1}{c}{SRoCC} & \multicolumn{1}{c}{KRoCC} & \multicolumn{1}{c|}{PLCC} & \multicolumn{1}{c}{SRoCC} & \multicolumn{1}{c}{KRoCC} & \multicolumn{1}{c|}{PLCC} & \multicolumn{1}{c}{SRoCC} & \multicolumn{1}{c}{KRoCC} & \multicolumn{1}{c}{PLCC} \\ \hline
\multicolumn{1}{l|}{\multirow{12}{*}{\begin{tabular}[c]{@{}l@{}}Perce\\-ption\end{tabular}}}  & brisque \cite{quality:brisque}                             & 0.1853                    & 0.1239                    & 0.2040                    & 0.2447                    & 0.1661                    & 0.2394                    & 0.2968                    & 0.2028                    & 0.3326                   \\
\multicolumn{1}{l|}{}                           & clipiqa \cite{quality:CLIPIQA}                             & 0.2134                    & 0.1427                    & 0.3343                    & 0.3961                    & 0.2688                    & 0.4996                    & 0.3500                    & 0.2379                    & 0.5254                   \\
\multicolumn{1}{l|}{}                           & cnniqa \cite{quality:CNNIQA}                              & 0.3397                    & 0.2309                    & 0.3910                    & 0.3429                    & 0.2338                    & 0.3806                    & 0.3178                    & 0.2168                    & 0.3857                   \\
\multicolumn{1}{l|}{}                           & dbcnn \cite{quality:DBCNN}                               & 0.4034                    & 0.2773                    & 0.4763                    & 0.4990                    & 0.3434                    & 0.5424                    & 0.4945                    & 0.3442                    & 0.5601                   \\
\multicolumn{1}{l|}{}                           & hyperiqa \cite{quality:HyperIQA}                            & 0.3352                    & 0.2295                    & 0.4268                    & 0.4510                    & 0.3069                    & 0.5071                    & 0.4318                    & 0.2946                    & 0.4984                   \\
\multicolumn{1}{l|}{}                           & liqe \cite{quality:LIQE}                                & 0.3431                    & 0.2333                    & 0.4029                    & 0.5061                    & 0.3455                    & 0.5395                    & 0.4966                    & 0.3417                    & 0.5664                   \\
\multicolumn{1}{l|}{}                           & musiq \cite{quality:musiq}                               & 0.3673                    & 0.2504                    & 0.4515                    & 0.5415                    & 0.3751                    & 0.6082                    & 0.5144                    & 0.3575                    & 0.5968                   \\
\multicolumn{1}{l|}{}                           & niqe \cite{quality:NIQE}                                & 0.0033                    & 0.0020                    & 0.0854                    & 0.0326                    & 0.0166                    & 0.1767                    & 0.1861                    & 0.1220                    & 0.3004                   \\
\multicolumn{1}{l|}{}                           & qalign \cite{quality:q-align}                              & \bf\textcolor{red}{0.6914}                    & \bf\textcolor{red}{0.4986}                    & \bf\textcolor{red}{0.6841}                    & \bf\textcolor{red}{0.7767}                    & \bf\textcolor{red}{0.5768}                    & \bf\textcolor{red}{0.7708}                    & \bf\textcolor{red}{0.7526}                    & \bf\textcolor{red}{0.5556}                    & \bf\textcolor{red}{0.7486}                   \\
\multicolumn{1}{l|}{}                           & topiq \cite{quality:topiq}                               & 0.4265                    & 0.2950                    & 0.4757                    & 0.5191                    & 0.3580                    & 0.5479                    & 0.4847                    & 0.3356                    & 0.5435                   \\
\multicolumn{1}{l|}{}                           & unique \cite{quality:unique}                              & 0.1389                    & 0.0940                    & 0.2346                    & 0.3198                    & 0.2134                    & 0.3987                    & 0.2944                    & 0.1980                    & 0.3991                   \\
\multicolumn{1}{l|}{}                           & wadiqam \cite{quality:wadiqam}                             & 0.2819                    & 0.1891                    & 0.3346                    & 0.2983                    & 0.2019                    & 0.3166                    & 0.2876                    & 0.1953                    & 0.3237                   \\ \hline
\multicolumn{1}{l|}{\multirow{4}{*}{\begin{tabular}[c]{@{}l@{}}Align\\-ment\end{tabular}}} & clip \cite{align:clip}                                & 0.3267                    & 0.2531                    & 0.3165                    & 0.4735                    & 0.3671                    & 0.5155                    & 0.4991                    & 0.3851                    & 0.5855                   \\
\multicolumn{1}{l|}{}                           & hps \cite{database/align:HPS}                                 & \bf\textcolor{blue}{0.6388}                    & \bf\textcolor{blue}{0.4583}                    & 0.4697                    & 0.6971                    & 0.5072                    & 0.6370                    & 0.6983                    & 0.5099                    & 0.5991                   \\
\multicolumn{1}{l|}{}                           & imagereward \cite{database/align:ImageReward}                         & 0.6247                    & 0.4452                    & \bf\textcolor{blue}{0.6421}                    & 0.6755                    & 0.4891                    & 0.7098                    & 0.6722                    & 0.4896                    & 0.7039                   \\
\multicolumn{1}{l|}{}                           & picscore \cite{database/align:PickAPic}                                 & 0.6372                    & 0.4566                    & 0.6368                    & \bf\textcolor{blue}{0.7090}                    & \bf\textcolor{blue}{0.5154}                    & \bf\textcolor{blue}{0.7124}                    & \bf\textcolor{blue}{0.7019}                    & \bf\textcolor{blue}{0.5109}                    & \bf\textcolor{blue}{0.7116} \\ \bottomrule                 
\end{tabular}
\end{table*}

Table \ref{tab:main} further lists the performance of 12 perceived qualities and 4 fit models on the sub-database of AIGIQA-20K. According to CFG, iterations are divided into normal (default) and abnormal (adjusted) resolutions. Overall, qalign remains the most accurate indicator of AIGC quality, despite ignoring the consistency of information between images and text. The three correlation indicators rank first in all sub-databases and lead the second by about 0.05. Among the other methods, picscore, imagereward, and hps, which take the T2I alignment into account, show a leading gap. Except for hps with certain defects in PLCC, all other models have acceptable performance and can be preliminarily used to predict the quality of AIGI. The zero-shot performance of other models is not ideal, and they must undergo fine-tuning similar to Table \ref{tab:train}, which limits their universality. Vertically comparing various sub-databases, we found that all models had more accurate AIGI evaluation results for the default CFG, but they performed better on abnormal data in terms of iteration times and resolution. Based on the error bar analysis in Figure \ref{fig:anal-para}, for the vast majority of T2I generative models, the limited iterations and compared to all iterations have a wider range of quality distribution compared to square resolution and rectangular resolution. Under larger quality differences, the accuracy of the evaluation will also further improve. As for CFG, there is no significant difference in the range of error bars. At this point, the more CFG deviates from the normal value, the more unnatural the generated results (such as AI artifacts, multi-finger content, etc.). Considering such distortion doesn't exist in NSIs, the zero-shot model alignment is not sensitive. Therefore, the more abnormal the CFG, the worse the performance of the evaluation.

\section{Conclusion}
\label{sec:conclusion}
In this paper, we establish the largest AIGI fine-grain quality database to date, AIGIQA-20K. We first select 15 mainstream T2I generation models and made dynamic adjustments on CFG, iteration, and resolution hyper-parameters for the first time. From this, 20,000 AIGIs are generated with different qualities to characterize the common images in today's AIGC community. Then, subjective quality labels are processed as the golden truth of quality. Finally, benchmark experiments are conducted to verify the performance of the current AIGI quality evaluator, including IQA and T2I alignment methods. Experimental results indicate that the universal zero-shot quality model is not yet complete and requires further development based on comprehensive subjective labels in this database.
% In summary, we believe this large-scale quality database can drive the evolution of AIGC for vision.
\vspace{-1mm}
\section*{Acknowledgment}
\vspace{-1mm}
The work was supported in part by the National Natural Science Foundation of China under Grant 62271312, 62301310; in part by the Shanghai Pujiang Program under Grant 22PJ1406800; in part by the China Postdoctoral Science Foundation under Grant 2023TQ0212; and in part by the Open Project of the Key Laboratory of Media Audio \& Video (Communication University of China), Ministry of Education.

%%%%%%%%% REFERENCES

{\small
\bibliographystyle{ieee_fullname}
\bibliography{egbib}
}

\end{document}